\documentclass[10pt,twocolumn,letterpaper]{article}

\usepackage[pagenumbers]{cvpr} %

\usepackage{amsmath}

\newcommand{\E}{\mathbb{E}}
\newcommand{\Ex}[2][]{\E_{#1}\left[{#2}\right]}

\usepackage{adjustbox}

\definecolor{cvprblue}{rgb}{0.21,0.49,0.74}
\usepackage[pagebackref,breaklinks,colorlinks,allcolors=cvprblue]{hyperref}

\def\paperID{21935} %
\def\confName{CVPR}
\def\confYear{2026}

\title{Intriguing Properties of Dynamic Sampling Networks}

\author{Dario Morle\textsuperscript{*}\\
McGill University\\
MILA-Quebec AI Institute\\
{\tt\small dario.morle@mail.mcgill.ca}
\and
Reid Zaffino\textsuperscript{*}\\
Independent Researcher\\
{\tt\small reid.zaffino@gmail.com}
}

\begin{document}
\maketitle
\footnotetext[1]{\textsuperscript{*}Co-first authors.}
\begin{abstract}

Dynamic sampling mechanisms in deep learning architectures have demonstrated utility across many computer vision models, though the theoretical analysis of these structures has not yet been unified. In this paper we connect the various dynamic sampling methods by developing and analyzing a novel operator which generalizes existing methods, which we term "warping".
Warping provides a minimal implementation of dynamic sampling which is amenable to analysis,
and can be used to reconstruct existing architectures including deformable convolutions, active convolutional units, and spatial transformer networks.
Using our formalism, we provide statistical analysis of the operator by modeling the inputs as both IID variables and homogeneous random fields.
Extending this analysis, we discover a unique asymmetry between the forward and backward pass of the model training.
We demonstrate that these mechanisms represent an entirely different class of orthogonal operators to the traditional translationally invariant operators defined by convolutions.
With a combination of theoretical analysis and empirical investigation, we find the conditions necessary to ensure stable training of dynamic sampling networks.
In addition, statistical analysis of discretization effects are studied.
Finally, we introduce a novel loss landscape visualization which utilizes gradient update information directly, to better understand learning behavior.

\end{abstract}

\section{Introduction}
\label{sec:intro}

Recent advancements in computer vision \cite{vaswani2017attention, 16x16words} have led to vision transformer architectures rising in prominence on a wide variety of tasks \cite{liu2021swin, touvron2021training, bao2021beit}. Much of this success is attributed to spatial attention \cite{zhu2019empirical}, though the exact contributions remain unclear. We attempt to demonstrate some intriguing properties of a specific instantiation of spatial attention, namely, dynamic sampling methods such as deformable convolutions \cite{defconv1}, spatial transformer networks \cite{stns}, and active convolutions \cite{acus}. Our methodology allows us to unify several previous architectures by demonstrating they are particular combinations of a component we call "warping" blocks, which are a simple form of dynamic sampling. Through our analysis, we demonstrate models which attempt to dynamically sample in this way have several intriguing properties, both experimentally and theoretically, and rectify some of the practical challenges they pose.

We define dynamic sampling as mechanisms which adaptively sample input data via interpolation methods. Previous work in dynamic sampling have been informed by effective receptive fields \cite{luo2016understanding}, and how traditional convolutional neural networks do not maximize usage of data effectively. The key insight from analysis of receptive fields has been the inefficiency of methods which rely solely on local short-ranged dependencies. Transformers have been shown to mitigate this issue by relating global information in image data, regardless of locality which was demonstrated in \cite{self-attention-vs-convs-2020}. The study of spatial attention in \cite{zhu2019empirical} shows that deformable convolutions act in the same manner as transformer based attention, with both utilizing query content and relative position.
Specific methods which use dynamic sampling are viewed as disparate methods, leading us to investigate and unify these methods both theoretically and practically. One notable gap we aim to address are the unique training dynamics these dynamic sampling structures impose, and the specific instability that arises when incorporating them into traditional deep learning training. For example, deformable convolution based networks often use pretraining and gradient coefficients on specific sub-networks to ensure training stability, which leads to significantly increased training time. The difficulty of training globally informed networks has also been studied in transformers as seen in \cite{liu2020understanding, xiong2020layer}, demonstrating these concerns are not limited to dynamic sampling methods. 

\subsection{Deformable Convolutions}
Deformable convolutions \cite{defconv1} are the most similar previous work to warping. These networks allow convolutional kernels to spatially vary their sampling points over the input in a data-dependent manner.  These networks were originally intended for more complex computer vision tasks such as object detection and segmentation, but also show improvements in classification. Deformable convolutions were further refined in \cite{defconv2}, which builds on the original paper by incorporating deformable convolutions throughout the entire network, rather than strictly later in the network.  They also add a modulation component to the deformable convolutions in place of traditional attention based methods. InternImage \cite{defconv3} was the first deformable convolution based architecture to achieve SOTA performance, and sets a strong performance benchmark similar to those made by vision transformers, by scaling up the previous work and incorporating more modern training methodologies. Further improvements to deformable convolutions were made in \cite{defconv4}, which optimizes the implementation of deformable convolutions and extends them to allow for an unbounded value range, thereby allowing the networks to better model long-range dependencies akin to transformer architectures. One slight variation of deformable convolutions are deformable kernels \cite{gao2019deformable}. These networks differ from the traditional deformable convolution as the kernels themselves are dynamically sampled, rather than the data they act on.

\subsection{Other Dynamic Sampling Methods}
Although deformable convolutions are the most prominent example of dynamic sampling networks in modern architectures, there are several other methods which have similar structures. Spatial transformer networks \cite{stns}, are one such example. The key idea of these spatial transformer networks are to allow the model to spatially manipulate the data being processed throughout the network, giving the model the ability to represent a wider set of geometric operations. Some of the most notable improvements to traditional convolutional neural networks provided by these methods include the ability to crop, scale, and rotate images. Dynamic filter networks \cite{jia2016dynamic} follow spatial transformer networks, but emphasize more local transformations, and generalize transformations to a greater set of filtering operations. Active convolutional units \cite{acus} are another prominent method that employs a unique sampling method for convolutions, though the offsets learned in ACUs are learned parameters, independent of the data they are operating on.

\subsection{Alternative Approaches to Spatial Attention}
Non-local nets \cite{nonlocalnets-2018} attempt to capture long-range dependencies by weighting features globally relative to a given position, in a similar manner to pairwise self attention. Global context networks \cite{global-context-nets-2020} expand on non-local nets by observing that the global contexts are agnostic to the query position. This fact is then leveraged in making global context blocks which more efficiently incorporate global information using a linear context vector as opposed to pairwise attention between pixels. CBAM (Convolutional Block Attention Module) \cite{CBAM} introduces lightweight modules for incorporating both channel and spatial attention into traditionally convolutional neural networks with little overhead. Gather and excitation networks \cite{hu2018gather} generalize the family of squeeze and excitation networks \cite{SENet} to be used for spatial context rather than channel-wise attention only. They achieve this by aggregating global spatial feature information (gathering), and redistributing this aggregated data to modulate the original signal (exciting). Another method for integrating attention and convolutions is dynamic convolutions \cite{chen2020dynamic}, which utilizes many predefined convolutional kernels and mixes them adaptively with attention to use a kernel conditioned on the input for the convolution operation. A somewhat similar method is employed in selective kernel networks \cite{li2019selective}, but the kernels combined are at different sizes to adjust the receptive field available to the models.

\section{Warping}
\label{sec:warping}

To analyze the properties of dynamic sampling, we define a transformation on a continuous field that we refer to as warping.  This transformation resamples a signal $x(t)$ using offsets $\varepsilon(t)$ to produce an output signal $y(t)$ as defined in equation \ref{eqn:py-cont}.

\begin{equation}
    y(t) = x(t + \varepsilon(t))
    \label{eqn:py-cont}
\end{equation}

To implement equation \ref{eqn:py-cont} in a neural network, the signals must be discretized.  We follow \cite{stns}, considering the signals to be constructed from discrete samples, ie. $x(t)=\sum_k x_k\varphi(t-k)$ for some interpolation kernel $\varphi(t)$.  We assume that $\varphi(0)=1$, $\varphi(-t)=\varphi(t)$, and that $\varphi$ is finitely integrable on $\mathbb{R}$.  Applying this discretization to the warping equation yields, 

\begin{equation}
    y_{n\mathbf{v}} = \sum_\mathbf{u}x_{n\mathbf{u}}\prod_{k}\varphi\left(\mathbf{v}_k-\mathbf{u}_k+\varepsilon_{k\mathbf{v}}\right)
    \label{eqn:py-disc}
\end{equation}

We denote the gradient of the loss $\mathcal{L}$ with respect to the output of a warping block $y$ as $\nabla_y\mathcal{L}$.  Using the chain rule, the gradient of $\mathcal{L}$ with respect to $x$ and $\varepsilon$ can be determined.  The details of this calculation can be found in the supplementary material.

\begin{equation}
    \nabla_x \mathcal{L}_{n\mathbf{w}} = \sum_\mathbf{v} \nabla_y \mathcal{L}_{n\mathbf{v}} \prod_k \varphi\left(\mathbf{v}_k - \mathbf{w}_k + \varepsilon_{k\mathbf{v}}\right)
    \label{eqn:gx-disc}
\end{equation}

\begin{multline}
    \nabla_\varepsilon \mathcal{L}_{i\mathbf{w}} =
    \sum_n \nabla_y \mathcal{L}_{n\mathbf{w}}\sum_\mathbf{u}x_{n\mathbf{u}}\\
    \varphi^\prime\left(\mathbf{w}_i-\mathbf{u}_i
    +\varepsilon_{i\mathbf{w}}\right)
    \prod_{k\neq i}\varphi\left(\mathbf{w}_k-\mathbf{u}_k+\varepsilon_{k\mathbf{w}}\right)
    \label{eqn:ge-disc}
\end{multline}

To demonstrate the generalization of warping, we provide mathematical formulations of various proposed models using warping.  Deformable convolutions \cite{defconv1} can be constructed using a concatenation of warping outputs with a 1x1 convolution.  For a deformable convolution with an offset grid $\mathcal{R}$ containing $N$ elements, we can define $y_{Nn+k}(t)=x_n(t+\varepsilon_k(t)+\mathcal{R}_k)$.  Performing a 1x1 convolution from $Nn$ channels to $n$ channels provides an implementation of the original deformable convolution proposal.  If the $\mathcal{R}_k$ term is considered as a bias applied to $\varepsilon_k(t)$, then this operation is $N$ warps of $x(t)$ that are concatenated along the channel dimension.  The remaining component needed to construct deformable convolutions from warping is that $\varepsilon_k(t)$ must be constructed from a convolution of $x(t)$.

Similar to deformable convolutions, it is possible to construct active convolutional units \cite{acus} from warping.  A similar setup for the output as was proposed for deformable convolutions can be used, with the exception that $\varepsilon_k$ is a constant that doesn't vary spatially.  Additionally, $\varepsilon_k$ would be directly trainable, rather than computed from $x(t)$.  While this results in some of our training stability analysis not applying to active convolutional units, much of our theoretical treatment of the transform still applies.

Spatial transformer networks \cite{stns} consider an arbitrary sub-network, termed a localization network, used to parameterize $\varepsilon$.  They primarily consider affine transforms in the image space, which can be represented using our framework by sampling from the vector field generated by the affine transform and using those samples to define $\varepsilon$.

\subsection{Applying Warping in Neural Networks}
\label{sec:warping.applied}

Using the definition provided in equation \ref{eqn:py-disc}, it is possible to construct a number of architectures using warping.  The most direct approach to this is to modify convolutions in existing architectures so that after the convolution, another convolutional layer is run to predict $\varepsilon$ from $x$ and uses the result to immediately warp $x$.  We term this architectural block SelfWarp.  Alternatively, for increased performance it is possible to jointly predict both $x$ and $\varepsilon$ in a single convolutional layer of $N+d$ output channels, where $N$ is the number of channels in $x$, and the remaining $d$ channels are used to parameterize $\varepsilon$.  We term this architectural block PackWarp, and while it can provide better training and inference performance, it is less amenable to analysis, so we instead consider the SelfWarp block for the remainder of this work.

Consider a convolutional deep learning architecture with skip connections modified to incorporate a dynamic sampling method, such as SelfWarp. If residual connections are used unmodified, spatial consistency will be broken as the manipulations performed by warping will not necessarily be aligned to the residual feature map. To avert this issue, the same $\varepsilon$ may be applied to the skip connection. 

\begin{figure}
    \centering
    \includegraphics[width=0.4\textwidth]{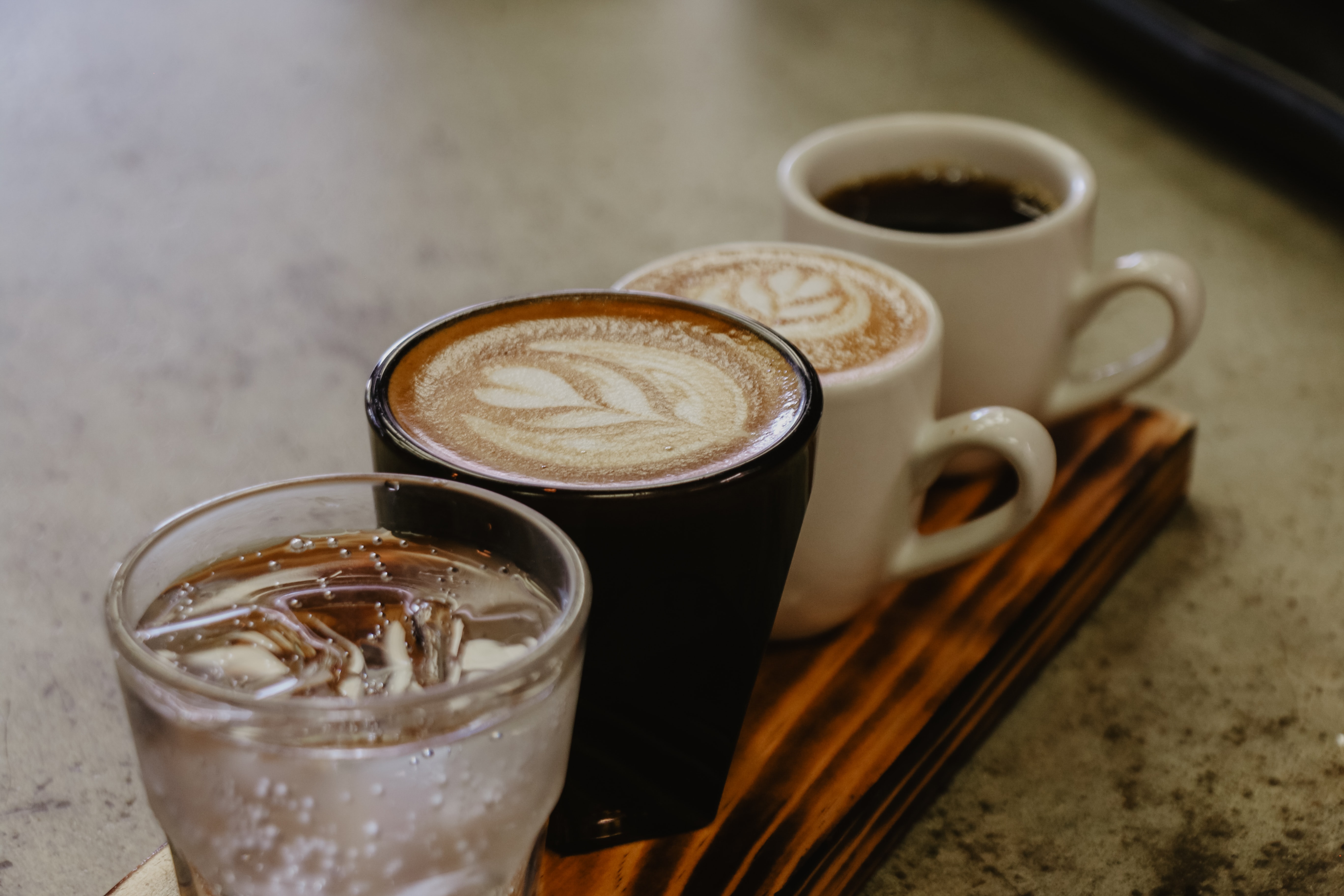}
    \includegraphics[width=0.4\textwidth]{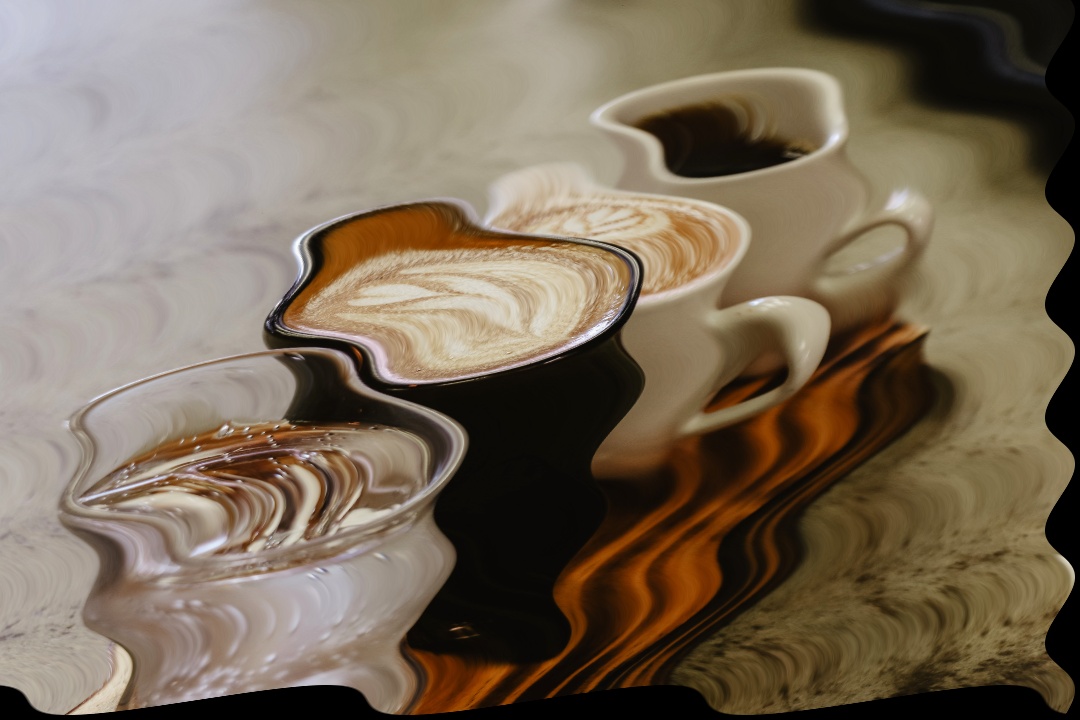}
    \caption{Comparison of an example input and warped image. The top image shows a default example, while the bottom displays a warped output of the same data.}
    \label{fig:warping-viz}
\end{figure}

One other design choice to note is the method of padding the image chosen when performing warping operations. Similar to standard convolutions, we opt to use zero padding, or equivalently, using black pixels to pad images. The practical implications of this occurs if a model attempts to sample from beyond the bounds of the image, leading to black edges being pulled into the warped data as seen in \ref{fig:warping-viz}. We consider this case to be undesirable for the network, as we suspect no useful information can be learned by the network beyond the bounds of the data itself.

To evaluate the contribution of a given warping layer, we use the average per-sample variance of $\varepsilon$. This allows us to measure whether the sampling points learned by the layer are uncorrelated, as correlated sampling directions would indicate the layer has merely learned to shift the data in an affine fashion, or was unable to learn any sampling better than an identity function. It should be noted that some utility could occur even with low variance of these layers, but we are attempting to encapsulate the unique benefits of using warping with this measure, rather than overall improvement to the model, which could be attributed to factors other than the inductive bias itself.

\section{Statistical Analysis}
\label{sec:stats}

Using the definition of warping from equation \ref{eqn:py-disc}, it is possible to statistically model the behavior of these blocks in a neural network.  If we consider $x$ and $\varepsilon$ to be composed of IID random variables, then the statistics of $y$ can be determined to follow equation \ref{eqn:py-iid}, where $d$ is the number of dimensions being warped over, and $\Phi_{a,b}$ is defined in equation \ref{eqn:phi-iid}.

\begin{equation}
    \begin{gathered}
    \mu_y = \mu_x\Phi_{1,1}^d\\
    \mu_y^2 + \sigma_y^2 = \mu_x^2\Phi_{1,2}^d + \sigma_x^2\Phi_{2,1}^d
    \end{gathered}
    \label{eqn:py-iid}
\end{equation}

\begin{equation}
    \begin{gathered}
    \Phi_{a,b} = \Ex{\left(\sum_n\varphi\left(\varepsilon-n\right)^a\right)^b}\\
    \Phi^\prime_{a,b} = \Ex{\left(\sum_n\varphi^\prime\left(\varepsilon-n\right)^a\right)^b}
    \end{gathered}
    \label{eqn:phi-iid}
\end{equation}

Similar analysis can be performed on the backwards pass to determine equations \ref{eqn:gx-iid} and \ref{eqn:ge-iid} for $\nabla_x\mathcal{L}$ and $\nabla_\varepsilon\mathcal{L}$, respectively.

\begin{equation}
    \begin{gathered}
    \mu_{\partial x}=\mu_{\partial y}\Phi_{1,1}^d\\
    \mu_{\partial x}^2+\sigma_{\partial x}^2=\mu_{\partial y}^2\Phi_{1,2}^d + \sigma_{\partial y}^2\Phi_{2,1}^d
    \end{gathered}
    \label{eqn:gx-iid}
\end{equation}

\begin{equation}
    \begin{gathered}
    \mu_{\partial\varepsilon}=N\mu_x\mu_{\partial y}\Phi_{1,1}^\prime\Phi_{1,1}^{d-1}\\
    \mu_{\partial\varepsilon}^2+\sigma_{\partial\varepsilon}^2=N\mu_x^2\left(N\mu_{\partial y}^2 + \sigma_{\partial y}^2\right)\Phi_{1,2}^\prime\Phi_{1,2}^{d-1}\\
        \hspace{60pt}+N\sigma_x^2\left(\mu_{\partial y}^2 + \sigma_{\partial y}^2\right)\Phi_{2,1}^\prime\Phi_{2,1}^{d-1}
    \end{gathered}
    \label{eqn:ge-iid}
\end{equation}

\subsection{Training Instability}
\label{sec:stats.stability}

To see the stability implications of this structure, we consider a SelfWarp block, where $\varepsilon$ is parameterized by $x$ through a convolution block, with $N$ input channels, and $d$ output channels.  For simplicity, we assume that both $x$ and the weight that parameterizes $\varepsilon$ have zero mean, with the weight having variance $\sigma_w^2$.  Assuming the convolutional kernel also consists of IID parameters, the variance equations from initialization theory \cite{glorot2010understanding} applies, and the variance of the update to the convolutional kernel $\sigma_{\partial w}^2$ can be shown to follow equation \ref{eqn:gw-iid}.

\begin{equation}
    \sigma_{\partial w}^2 = dNk^2\sigma_x^4\sigma_{\partial y}^2\Phi_{2,1}^\prime\Phi_{2,1}^{d-1}
    \label{eqn:gw-iid}
\end{equation}

Equation \ref{eqn:gw-iid} shows that the variance of the update depends quadratically on the forward variance $\sigma_x^2$.  This quadratic dependence on the forward variance empirically results in unstable training dynamics, causing the variance of the convolutional kernel to explode within a few training iterations.  This dependency can be avoided by inserting an additional normalization layer between $x$ and $\varepsilon$, thereby separating the forward variances, removing one of the $\sigma_x^2$ terms from equation \ref{eqn:gw-iid}.

In addition to the instabilities caused by the quadratic dependence demonstrated in equation \ref{eqn:gw-iid}, we find that dynamic sampling methods are naturally susceptible to exploding gradients.  Traditionally, a condition required to avoid the exploding or vanishing gradient problems during training is that on average, when layers are characterized as functions, the determinant of their Jacobian is approximately 1 \cite{dynamical-isometry}.  A direct implication of this is that when applied to IID normal random variables, the coefficient on the forward variance needs to match the coefficient on the backward variance, ie $\sigma_y/\sigma_{\partial x}\approx \sigma_x/\sigma_{\partial y}$.  While an initial inspection of equations \ref{eqn:py-iid} and \ref{eqn:gx-iid} would suggest that this condition is met, these equations consider the inputs and outputs of a warping block to be disconnected from each other.  Consider the case of SelfWarp again where $\varepsilon$ is parameterized by $x$ through a convolutional block.  In this case, the updates into $x$ would consist of the sum of the updates from equation \ref{eqn:gx-disc} and equation \ref{eqn:ge-disc} with the appropriate coefficient implied by the sub-network parameterizing $\varepsilon$.  The updates coming from this branch will cause each warping block's backward variance to be elevated, while the forward variance will remain mostly unchanged, causing exploding gradients.  Note that as $\sigma_\varepsilon$ increases, $\Phi_{a,b}$ will be mostly unaffected, see the supplementary material for a precise characterization of $\Phi_{a,b}$.

To mitigate this source of instability, one would need to ensure that the gradients coming from the $\varepsilon$ path into $x$ are sufficiently small as to not cause a fast exponential growth in the variance of the gradients.  Empirically, we find that this can be accomplished through a variety of methods.  The most direct is to ignore the gradients coming from the $\varepsilon$ branch.  Alternatively, initializing the variance of $\varepsilon$ to be sufficiently small prevents this source of instability.

Due to there being multiple sources of instability in warping, mechanisms used to address each must be applied simultaneously to ensure stable training.  Alleviating only the quadratic dependency, or only preventing exploding gradients would not stabilize training. Both of these sources of instability can also be mitigated by starting with an initialization structure where $\sigma_x^2$ is small as to limit the updates into the sub-network and having a sufficiently small learning rate on the sub-network, although this approach appears to be sensitive to both model depth and initialization.  This was the approach used by deformable convolutions \cite{defconv1}, where the initialization structure was generated by pre-training the neural network with all $\varepsilon=0$.

Since both of these sources of instability only effects the $\varepsilon$ branch of a warping block, we do not expect gradient problems to manifest in it's typical form.  We obverse that the $\varepsilon$ branches alone undergo an exploding gradient process, but through training iterations rather than sequential layers. If the warping block is implemented by assuming zero padding outside the range of $x(t)$, then $\varepsilon$ will eventually sample points entirely outside the boundary, causing the network to collapse into all zeros in the forward pass.  Eventually, weight decay will bring $\varepsilon$ back into a range where it is sampling from $x(t)$, and this cycle will repeat.  Alternatively, if the warping block is implemented such that $x(t)$ is cyclic as assumed in the correlational analysis shown in section \ref{sec:cont}, $\varepsilon$ will simply diverge to infinity.

\section{Continuous Analysis}
\label{sec:cont}

To extend the statistical analysis presented in the previous section, we also consider warping as a linear transform.  In general, we formulate linear layers of a neural network as an integral transform of the input function $x(t)$,

\begin{equation}
    y(t) = \int_{\mathbb{R}^N}K(t, \tau)x(\tau)d\tau
    \label{eqn:int-tfm-fwd}
\end{equation}

Throughout this section, we use $N$ rather than $d$ to denote the number of dimensions being operated on, to prevent notational ambiguity.  Convolutions are a special case of this transform where the integral kernel $K$ can be represented as a convolutional kernel $K(t, \tau) = h(t - \tau)$.  In this work, we explore the kernel structure of $K(t, \tau) = \delta(\tau - f(t))$ with offsets $f(t)=t+\varepsilon(t) : \mathbb{R}^N \rightarrow \mathbb{R}^N$.  Substituting this definition into equation \ref{eqn:int-tfm-fwd} results in the continuous form of warping proposed in section \ref{sec:warping}: $y(t) = x(t+\varepsilon(t))$. From the general form provided in equation \ref{eqn:int-tfm-fwd}, the gradient update to the input signal $x(t)$ can de determined by taking the functional derivative through $y(t)$, resulting in the following update transformation,

\begin{equation}
    \begin{split}
    \nabla_x\mathcal{L}(t)
    &= \frac{\delta\mathcal{L}}{\delta x(t)} \\
    &= \int_{\mathbb{R}^N}\frac{\delta\mathcal{L}}{\delta y(\tau)}\frac{\delta y(\tau)}{\delta x(t)}d\tau \\
    &= \int_{\mathbb{R}^N}\nabla_y\mathcal{L}(\tau)\frac{\delta}{\delta x(t)}\int_{\mathbb{R}^N}K(\tau, s)x(s)dsd\tau \\
    &= \int_{\mathbb{R}^N}\nabla_y\mathcal{L}(\tau)\int_{\mathbb{R}^N}K(\tau, s)\frac{\delta x(s)}{\delta x(t)}dsd\tau \\
    &= \int_{\mathbb{R}^N}\nabla_y\mathcal{L}(\tau)\int_{\mathbb{R}^N}K(\tau, s)\delta(s - t)dsd\tau \\
    &= \int_{\mathbb{R}^N}K(\tau, t)\nabla_y\mathcal{L}(\tau)d\tau \\
    \end{split}
    \label{eqn:int-tfm-bwd}
\end{equation}

This transformation to the gradient update is the integral transform corresponding to the transpose of the kernel $K$ of the forward pass.  Using the integral transform kernel definition for convolutions allows us to easily determine the resulting transpose of the kernel: $K^T(t, \tau) = h(\tau - t)$.  This result matches the discrete implementation of gradients through convolutional layers where the kernel is spatially flipped in the backwards pass.  In the case of warping, a direct application of the transpose results in the following transformation kernel: $K^T(t, \tau) = \delta(t - f(\tau))$.  Assuming $f(t)$ is a bijective continuous function, then,

\begin{equation}
    K^T(t, \tau) = \frac{\delta(\tau - f^{-1}(t))}{\left|\mathbf{J_f}(f^{-1}(t))\right|}
    \label{eqn:warping-bwd-kernel}
\end{equation}

\subsection{Covariance Analysis}

Using the continuous representation of warping, it is possible to understand the effects of warping on a broader class of input signals and generalize the analysis presented in section \ref{sec:stats}.  Consider the case where $x(t)$ and $\varepsilon(t)$ are homogeneous random fields with autocorrelation functions $R_x(t)$ and $R_\varepsilon(t)$, respectively.  Note that we are considering the components of $\varepsilon(t)$ to be independent, all with autocorrelation $R_\varepsilon(t)$.  The power spectral density $S_y(\xi)$ can be shown to follow equation \ref{eqn:psd_y}, where the input power spectral density is $S_x(\xi)$, the normalized autocorrelation of the random field $\varepsilon(t)$ is $\rho_\varepsilon(t)$, and $r=4\pi^2\sigma_\varepsilon^2||\xi-\alpha||^2$.  The derivation of this equation can be found in the supplementary material.

\begin{multline}
    S_y(\xi) =\\\sum_{\alpha\in\mathbb{Z}^N}S_x(\xi-\alpha)
        \int_{[0,1]^N}e^{-r\left(1-\rho_\varepsilon(t)\right)}e^{-2\pi i\alpha\cdot t}dt
    \label{eqn:psd_y}
\end{multline}

Extending this analysis to the backwards pass at first seems to require a separate derivation akin to the analysis presented in the supplementary material, but using the kernel representation of warping, both backwards passes can be written in terms of the kernel $K(t,\tau)=\delta(\tau-f(t))$.

\subsection{Warping as an Orthogonal Transform}
\label{sec:cont.ortho}

Consider the transformation defined by first applying a warping operation $K$, then applying the backwards pass $K^T$, both with the same offsets $f(t)$.  The overall transformation can be evaluated as the product $KK^T$ as shown in equation \ref{eqn:fwd-bwd-kernel}.

\begin{equation}
    \begin{split}
    KK^T(t,\tau)
    &=\int_{\mathbb{R}^N}K(t,s)K^T(s,\tau)ds\\
    &=\int_{\mathbb{R}^N}\delta(s-f(t))\frac{\delta(\tau-f^{-1}(s))}{|\mathbf{J_f}(f^{-1}(s))|}ds\\
    &=\frac{\delta(\tau-t)}{|\mathbf{J_f}(t)|}
    \end{split}
    \label{eqn:fwd-bwd-kernel}
\end{equation}

The resulting transform from equation \ref{eqn:fwd-bwd-kernel} is simply the identity transform, scaled by the Jacobian of the offsets $f(t)$.  This establishes the backwards pass as a pseudo-inverse of the forward pass.  Additionally, it also classifies warping as a type of orthogonal transform.  This is categorically different than the class of translationally invariant operators defined by convolutions.

\subsection{Generalization of IID Characteristics}

To demonstrate the generality provided by this correlational approach, we reconstruct the zero mean case of the IID equations provided in section \ref{sec:stats}.  The primary challenge in performing this reconstruction is capturing the discretization effects of the interpolation kernel $\varphi(t)$.   These effects can be captured by considering the signal $x(t)$ to be constructed from a convolution with $\varphi(t)$, ie. $x(t)\to x(t)*\varphi(t)$.  Applying this to equation \ref{eqn:psd_y} and evaluating for $\sigma_y^2=R_y(0)$ yields the zero mean case of equation \ref{eqn:py-iid} for any choice of $R_\varepsilon(t)$ and $R_x(t)$.  Details of this analysis can be found in the supplementary material.  Due to the assumption of a zero-mean random field in the derivation of equation \ref{eqn:psd_y}, it is not possible to fully recover the IID equations presented in section \ref{sec:stats}.

This result demonstrates that the characteristics derived from the IID analysis presented in section \ref{sec:stats.stability} generalize beyond the initial phase of training where the IID assumption is broadly correct.  Although the exploding gradient problem can be mitigated with pretraining, the quadratic dependence instability that arises from the statistical structure of dynamic sampling is nonetheless subject to the implications of section \ref{sec:stats.stability}.

\section{Results}
\label{sec:results}

To demonstrate the stability conditions identified in section \ref{sec:stats.stability}, we consider a Resnet model with all convolutional layers replaced by SelfWarp layers.  We also consider the case where skip connections are warped using the $\varepsilon$ values predictions from the residual blocks to ensure spatial consistency as described in section \ref{sec:warping.applied}.  As described in section \ref{sec:stats.stability}, two sources of instability must be mitigated to ensure stable training.  To demonstrate this, we ablate one method for removing the quadratic dependency of the $\varepsilon$ convolutional kernel on $x$, and two different methods to remove the exploding gradients caused by the unbalanced gradient variance from the $\varepsilon$ path.  To remove the quadratic dependence, we insert a batchnorm block between $x$ and the convolution parameterizing $\varepsilon$. As for exploding gradients, we either ignore the gradients from $\varepsilon$ when computing gradients into $x$ by detaching $x$ from the backwards graph when computing $\varepsilon$.  Or we ensure that the backward variance from $\varepsilon$ is sufficiently small by initializing all batchnorm multiplication terms to $0.1$, and rely on training dynamics to prevent this value from becoming too large during training.  We run all training variants using a Resnet 56 model \cite{Resnet} on Cifar 10 \cite{cifar}, using an identical training scheme to Resnet.  The results of these training runs are summarized in table \ref{tab:stability-ablations}.  For each of these trials, we consider a trial to have diverged, if during the course of training, $\varepsilon$ significantly exceeds the expected range.  In particular, we consider a training run to have diverged if there exists a layer for which $\sigma_{\varepsilon}^2>1000$ at some point during training.  Note that it will often be the case that eventually these divergent trials will return to a normal range after a sufficient number of epochs due to the effects described in section \ref{sec:stats.stability}.

\begin{table}
    \centering
    \adjustbox{max width=\columnwidth}{
    \begin{tabular}{cccc}
    \hline
        \textbf{Pre-$\varepsilon$ BN} &
        \textbf{\begin{tabular}{c}
            Batch Norm \\
            Initialization
        \end{tabular}}
        &
        \textbf{\begin{tabular}{c}
            Warp \\
            Grad Type
        \end{tabular}}
        &
        \textbf{\begin{tabular}{c}
            Result/ \\
            Accuracy
        \end{tabular}}\\
    \hline
    No  & 1.0 & full grads & diverges \\
    No  & 1.0 & detached   & diverges \\
    No  & 0.1 & full grads & diverges \\
    No  & 0.1 & detached   & diverges \\
    Yes & 1.0 & full grads & diverges \\
    Yes & 1.0 & detached   & 92.378\% \\
    Yes & 0.1 & full grads & \textbf{93.620\%} \\
    Yes & 0.1 & detached   & 93.209\% \\
    \hline
    \end{tabular}
    }
    \caption{Experiment results for different normalization, initialization, and gradient configurations.}
    \label{tab:stability-ablations}
\end{table}

As is demonstrated in table \ref{tab:stability-ablations}, to achieve stable training of these models, both a mechanism for mitigating the quadratic dependence and a mechanism for mitigating the exploding gradients are necessary to stabilize training. Both variants which utilized ignoring gradients from $\varepsilon$ to prevent exploding gradients underperformed the alternative method of mitigation.  This is likely due to these gradients, although destabilizing, still providing useful information which the remainder of the model could adapt to.  As demonstrated in section \ref{sec:cont.ortho}, dynamic sampling parameterizes a categorically different set of operators than convolutions which likely require a different representation that the model would otherwise be unable to adapt to.

In a Resnet \cite{Resnet} architecture using SelfWarps, the parameter count increases by 4.7\%.  It should be noted that warping requires approximately twice the training time relative to the base Resnet architecture. In comparison, converting all layers of a Resnet to deformable convolutions \cite{defconv1} increases the parameter count by 46.2\%, and will typically require around 6x the training time of a base Resnet. Though, the increased training cost and parameter count proportionally decreases with wider architectures.

To demonstrate the extension of our analysis to other methods of dynamic sampling, we apply our stability conditions to deformable convolutions.  Instead two mechanisms for mitigating exploding gradients, we only consider the batchnorm initialization approach as it yields better results.  We also consider the training approach used by deformable convolutions by allowing for both pretraining and gradient scaling on the parameters in the $\varepsilon$ branch.  The results of these trials are shown in table \ref{tab:defconv-ablation}, where the diverges label has the same meaning as in table \ref{tab:stability-ablations}.
 
\begin{table}
    \centering
    \adjustbox{max width=\columnwidth}{
    \begin{tabular}{ccccc}
    \hline
        \textbf{Pretraining} &
        \textbf{\begin{tabular}{c}
            Gradient \\
            Scaling
        \end{tabular}} &
        \textbf{\begin{tabular}{c}
                Batch Norm \\
                Initialization
            \end{tabular}} &
        \textbf{Pre-$\varepsilon$ BN} &
        \textbf{\begin{tabular}{c}
            Result/ \\
            Accuracy
        \end{tabular}}\\
    \hline

    Yes & No  & 1.0 & No  & diverges \\ %
    Yes & Yes & 1.0 & No  & \textbf{93.760\%} \\ %
    No  & No  & 0.1 & No  & diverges \\ %
    No  & No  & 1.0 & No  & diverges \\ %
    No  & Yes & 0.1 & No  & 92.448\% \\ %
    No  & Yes & 1.0 & No  & 92.228\% \\ %
    No  & No  & 0.1 & Yes & 92.648\% \\ %
    
    \hline
    \end{tabular}
    }
    \caption{Results for varying pretraining, deformable convolution gradient scaling, batch normalization initialization coefficients, and adding pre-$\varepsilon$ batch Norm configurations for deformable convolutions implemented with warping.}
    \label{tab:defconv-ablation}
\end{table}

To evaluate the importance of spatial consistency, we consider the best model from table \ref{tab:stability-ablations} and ablate the application of the warping transform to the skip connection.  Without applying the warping to the skip connection, the accuracy of the resulting model drops to 92.919\%, significantly lower than the 93.620\% achieved by applying the warping transforms to the skip connection.  These results demonstrate that spatial consistency does appear to be a relevant property for model accuracy.

\subsection{Bifurcation of Layer Variances}

\begin{figure}
    \centering
    \includegraphics[width=0.5\textwidth]{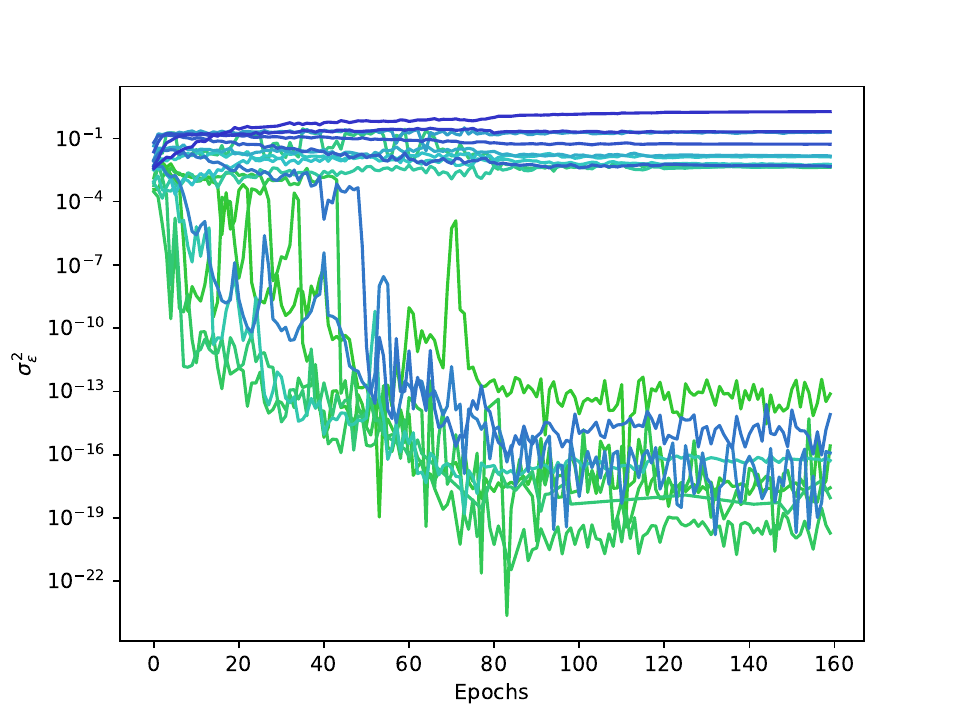}
    \caption{Training a SelfWarp Resnet 20 on Cifar 10 \cite{cifar}.  Early layers are shown in green (light), later layers are shown in blue (dark).}
    \label{fig:bifurcation}
\end{figure}

As stated in \ref{sec:warping.applied}, we use the average per-sample variance of $\varepsilon$ to measure a given warping layer's impact. When training is stable, we notice that layer variances will either converge to 0, or remain elevated throughout training, after an initial phase of growth.  A typical training run where this occurs is shown in figure \ref{fig:bifurcation}.

In these training runs, the smallest variances of the higher variance group will typically be sufficiently small such that the effect of warping is negligible.  This prevents a simple explanation such as weight decay having a dominating effect on layers where the mapping is trivial, and gradients maintaining a higher variance on layers where the representation is useful.

Additionally, this bifurcation of layer variances tends to select for later layers maintaining higher variances, while layers close to the input are more likely to tend towards zero variance.  The exact number of layers in each group, and the choice of which layer numbers settle in each group varies between training runs.  This behavior seems to be consistent across models architecture choices, and model depth. We also note that this bifurcation phenomena is also observed across model sizes and in deformable convolution networks.

\subsection{Loss Landscape}

Following \cite{li2018visualizing}, we visualize the loss landscape of warping networks. The original loss landscape is generated by performing filter-wise normalization, to remove the potential scaling effects contributed by model weight sizes. We encounter a different issue when trying to use these methods for dynamic sampling networks, though we believe our solution is also suitable as a general purpose method for generating loss landscapes. Since the convolution weights in the $\varepsilon$ branches of SelfWarps are initialized to be 0, the traditional approach of normalizing the scaling directions would cause the visualization to simply ignore these layers. Rather than computing directions based on the parameter weight sizes and filtering by the entire layer, we aim to parameterize the directions of the loss landscape by utilizing gradient update information instead. This more accurately reflects the topology nearby a given model in a training environment. Our method for obtaining these gradient-informed directions of a network with parameters $\theta$ starts with generating a gaussian random direction vector $\text{d}$, indexed by $i$ and $j$, where $i$ is the layer in the network, and $j$ corresponds to index of the parameter within a layer. Next, the vector is normalized and multiplied by the norm of the expected value of the weights gradient with respect to layer $i$, as seen in \ref{eqn:grad_noise_rescale}. 

\begin{equation}
    \begin{gathered}
    \mathbf{d}_{i,j} \sim \mathcal{N}(0, 1), \\
    \mathbf{d}_{i,j} \leftarrow 
    \mathbf{d}_{i,j}
    \frac{\lVert\Ex[x]{\nabla_{\theta_i}\mathcal{L}}\rVert}{\lVert\mathbf{d}_{i}\rVert}
    \end{gathered}
    \label{eqn:grad_noise_rescale}
\end{equation}

\begin{figure}
    \centering
    \includegraphics[width=0.4\textwidth]{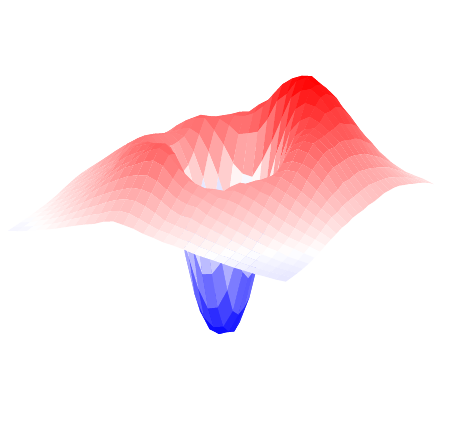}
    \caption{Loss landscape of a Resnet-56 based SelfWarp model shown jointly over all model parameters.}
    \label{fig:lossland}
\end{figure}

\begin{figure}
    \centering
    \includegraphics[width=0.4\textwidth]{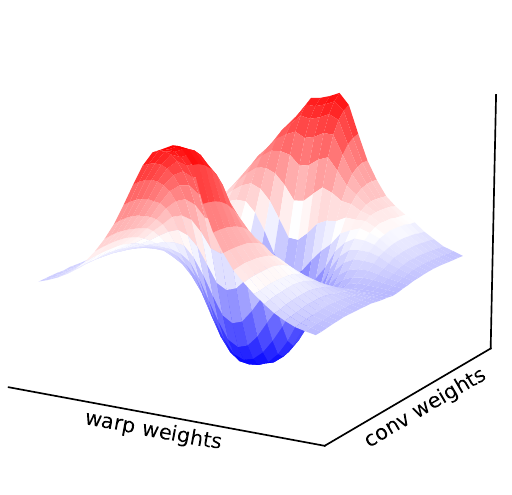}
    \caption{Loss landscape of a Resnet-56 based SelfWarp model with separated warping and non-warping parameters.}
    \label{fig:warpland}
\end{figure}

From this method we are able to generate two distinct plots to evaluate the qualitative characteristics of dynamic sampling. The first loss landscape \ref{fig:lossland} is generated with each direction vector sampling from all convolutions and parameters from the $\varepsilon$ branch equally, leading to a landscape similar to traditional Resnet landscape visualizations. This is contrasted by the second landscape \ref{fig:warpland}, where each axis varies independently over each of the weight types specified prior. This demonstrates the different training characteristics of the two operations, where we see that warping parameters plateau loss given a large enough distance from the fully trained center, whereas convolutional parameters contribute more loss as they stray further from the initially trained model. This is due to large $\varepsilon$ causing the warping blocks to sample out of the spatial bounds of the input, causing the output to become zero for all warping layers.  The result is the last layer predicting a uniform distribution over all classes, saturating the loss.  We compute these loss landscapes on the best model from table \ref{tab:stability-ablations}, resulting in figure \ref{fig:lossland} for the joint direction vectors, and figure \ref{fig:warpland} for the disentangled directions.

\section{Conclusion}
\label{sec:conc}

In this work, we analyze the properties of dynamic sampling in neural networks through the definition of warping.  Warping provides a minimal implementation of dynamic sampling from which existing work can be constructed, allowing our analysis to generalize.  Through a statistical analysis of warping, we find two sources of instability intrinsic to the structure of dynamic sampling and provide mechanisms to stabilize training.  We extend our analysis by consider the effects of warping on homogeneous random fields, deriving a generalized form of our statistical analysis which strengthens our stability claims.  In this analysis, we also show that warping parameterizes a set of orthogonal transforms, providing a categorically different representation than the translationally invariant transforms parameterized by convolutions.  Using our analysis, we demonstrate the validity of our stability conditions through ablation studies.

A potential avenue for improvement could be further study of the bifurcation effect we observed, determining if other architectures display similar behavior, or if the phenomenon is intrinsic to dynamic sampling networks. Additionally, integrating warping into more modern architectures and testing with larger-scale models and datasets would provide additional empirical validation of the trends we have investigated.

{
    \small
    \bibliographystyle{ieeenat_fullname}
    \bibliography{main}

@String(ECCV= {Eur. Conf. Comput. Vis.})

@String(ECCV  = {ECCV})

@inproceedings{nonlocalnets-2018,
  title={Non-local neural networks},
  author={Wang, Xiaolong and Girshick, Ross and Gupta, Abhinav and He, Kaiming},
  booktitle={Proceedings of the IEEE conference on computer vision and pattern recognition},
  pages={7794--7803},
  year={2018}
}

@article{global-context-nets-2020,
  title={Global context networks},
  author={Cao, Yue and Xu, Jiarui and Lin, Stephen and Wei, Fangyun and Hu, Han},
  journal={IEEE Transactions on Pattern Analysis and Machine Intelligence},
  volume={45},
  number={6},
  pages={6881--6895},
  year={2020},
  publisher={IEEE}
}

@inproceedings{CBAM,
  title={Cbam: Convolutional block attention module},
  author={Woo, Sanghyun and Park, Jongchan and Lee, Joon-Young and Kweon, In So},
  booktitle={Proceedings of the European conference on computer vision (ECCV)},
  pages={3--19},
  year={2018}
}

@inproceedings{SENet,
  title={Squeeze-and-excitation networks},
  author={Hu, Jie and Shen, Li and Sun, Gang},
  booktitle={Proceedings of the IEEE conference on computer vision and pattern recognition},
  pages={7132--7141},
  year={2018}
}

@inproceedings{Resnet,
  title={Deep residual learning for image recognition},
  author={He, Kaiming and Zhang, Xiangyu and Ren, Shaoqing and Sun, Jian},
  booktitle={Proceedings of the IEEE conference on computer vision and pattern recognition},
  pages={770--778},
  year={2016}
}

@misc{self-attention-vs-convs-2020,
      title={On the Relationship between Self-Attention and Convolutional Layers}, 
      author={Jean-Baptiste Cordonnier and Andreas Loukas and Martin Jaggi},
      year={2020},
      eprint={1911.03584},
      archivePrefix={arXiv},
      primaryClass={cs.LG},
      url={https://arxiv.org/abs/1911.03584}, 
}

@inproceedings{16x16words,
    title={An Image is Worth 16x16 Words: Transformers for Image Recognition at Scale},
    author={Alexey Dosovitskiy and Lucas Beyer and Alexander Kolesnikov and Dirk Weissenborn and Xiaohua Zhai and Thomas Unterthiner and Mostafa Dehghani and Matthias Minderer and Georg Heigold and Sylvain Gelly and Jakob Uszkoreit and Neil Houlsby},
    booktitle={International Conference on Learning Representations},
    year={2021},
    url={https://openreview.net/forum?id=YicbFdNTTy}
}

@article{stns,
  title={Spatial transformer networks},
  author={Jaderberg, Max and Simonyan, Karen and Zisserman, Andrew and others},
  journal={Advances in neural information processing systems},
  volume={28},
  year={2015}
}

@inproceedings{defconv1,
  title={Deformable convolutional networks},
  author={Dai, Jifeng and Qi, Haozhi and Xiong, Yuwen and Li, Yi and Zhang, Guodong and Hu, Han and Wei, Yichen},
  booktitle={Proceedings of the IEEE international conference on computer vision},
  pages={764--773},
  year={2017}
}

@inproceedings{defconv2,
  title={Deformable convnets v2: More deformable, better results},
  author={Zhu, Xizhou and Hu, Han and Lin, Stephen and Dai, Jifeng},
  booktitle={Proceedings of the IEEE/CVF conference on computer vision and pattern recognition},
  pages={9308--9316},
  year={2019}
}

@inproceedings{defconv3,
  title={Internimage: Exploring large-scale vision foundation models with deformable convolutions},
  author={Wang, Wenhai and Dai, Jifeng and Chen, Zhe and Huang, Zhenhang and Li, Zhiqi and Zhu, Xizhou and Hu, Xiaowei and Lu, Tong and Lu, Lewei and Li, Hongsheng and others},
  booktitle={Proceedings of the IEEE/CVF conference on computer vision and pattern recognition},
  pages={14408--14419},
  year={2023}
}

@inproceedings{defconv4,
  title={Efficient deformable convnets: Rethinking dynamic and sparse operator for vision applications},
  author={Xiong, Yuwen and Li, Zhiqi and Chen, Yuntao and Wang, Feng and Zhu, Xizhou and Luo, Jiapeng and Wang, Wenhai and Lu, Tong and Li, Hongsheng and Qiao, Yu and others},
  booktitle={Proceedings of the IEEE/CVF Conference on Computer Vision and Pattern Recognition},
  pages={5652--5661},
  year={2024}
}

@inproceedings{dynamical-isometry,
  title={Dynamical isometry and a mean field theory of cnns: How to train 10,000-layer vanilla convolutional neural networks},
  author={Xiao, Lechao and Bahri, Yasaman and Sohl-Dickstein, Jascha and Schoenholz, Samuel and Pennington, Jeffrey},
  booktitle={International conference on machine learning},
  pages={5393--5402},
  year={2018},
  organization={PMLR}
}

@inproceedings{chen2020dynamic,
  title={Dynamic convolution: Attention over convolution kernels},
  author={Chen, Yinpeng and Dai, Xiyang and Liu, Mengchen and Chen, Dongdong and Yuan, Lu and Liu, Zicheng},
  booktitle={Proceedings of the IEEE/CVF conference on computer vision and pattern recognition},
  pages={11030--11039},
  year={2020}
}

@article{hu2018gather,
  title={Gather-excite: Exploiting feature context in convolutional neural networks},
  author={Hu, Jie and Shen, Li and Albanie, Samuel and Sun, Gang and Vedaldi, Andrea},
  journal={Advances in neural information processing systems},
  volume={31},
  year={2018}
}

@inproceedings{acus,
  title={Active convolution: Learning the shape of convolution for image classification},
  author={Jeon, Yunho and Kim, Junmo},
  booktitle={Proceedings of the IEEE conference on computer vision and pattern recognition},
  pages={4201--4209},
  year={2017}
}

@inproceedings{li2019selective,
  title={Selective kernel networks},
  author={Li, Xiang and Wang, Wenhai and Hu, Xiaolin and Yang, Jian},
  booktitle={Proceedings of the IEEE/CVF conference on computer vision and pattern recognition},
  pages={510--519},
  year={2019}
}

@article{vaswani2017attention,
  title={Attention is all you need},
  author={Vaswani, Ashish and Shazeer, Noam and Parmar, Niki and Uszkoreit, Jakob and Jones, Llion and Gomez, Aidan N and Kaiser, {\L}ukasz and Polosukhin, Illia},
  journal={Advances in neural information processing systems},
  volume={30},
  year={2017}
}

@article{luo2016understanding,
  title={Understanding the effective receptive field in deep convolutional neural networks},
  author={Luo, Wenjie and Li, Yujia and Urtasun, Raquel and Zemel, Richard},
  journal={Advances in neural information processing systems},
  volume={29},
  year={2016}
}

@inproceedings{zhu2019empirical,
  title={An empirical study of spatial attention mechanisms in deep networks},
  author={Zhu, Xizhou and Cheng, Dazhi and Zhang, Zheng and Lin, Stephen and Dai, Jifeng},
  booktitle={Proceedings of the IEEE/CVF international conference on computer vision},
  pages={6688--6697},
  year={2019}
}

@article{gao2019deformable,
  title={Deformable kernels: Adapting effective receptive fields for object deformation},
  author={Gao, Hang and Zhu, Xizhou and Lin, Steve and Dai, Jifeng},
  journal={arXiv preprint arXiv:1910.02940},
  year={2019}
}

@article{jia2016dynamic,
  title={Dynamic filter networks},
  author={Jia, Xu and De Brabandere, Bert and Tuytelaars, Tinne and Gool, Luc V},
  journal={Advances in neural information processing systems},
  volume={29},
  year={2016}
}

@misc{cifar,
  title={Learning multiple layers of features from tiny images.(2009)},
  author={Krizhevsky, Alex and Hinton, Geoffrey and others},
  year={2009}
}

@article{li2018visualizing,
  title={Visualizing the loss landscape of neural nets},
  author={Li, Hao and Xu, Zheng and Taylor, Gavin and Studer, Christoph and Goldstein, Tom},
  journal={Advances in neural information processing systems},
  volume={31},
  year={2018}
}

@inproceedings{liu2021swin,
  title={Swin transformer: Hierarchical vision transformer using shifted windows},
  author={Liu, Ze and Lin, Yutong and Cao, Yue and Hu, Han and Wei, Yixuan and Zhang, Zheng and Lin, Stephen and Guo, Baining},
  booktitle={Proceedings of the IEEE/CVF international conference on computer vision},
  pages={10012--10022},
  year={2021}
}

@inproceedings{touvron2021training,
  title={Training data-efficient image transformers \& distillation through attention},
  author={Touvron, Hugo and Cord, Matthieu and Douze, Matthijs and Massa, Francisco and Sablayrolles, Alexandre and J{\'e}gou, Herv{\'e}},
  booktitle={International conference on machine learning},
  pages={10347--10357},
  year={2021},
  organization={PMLR}
}

@article{bao2021beit,
  title={Beit: Bert pre-training of image transformers},
  author={Bao, Hangbo and Dong, Li and Piao, Songhao and Wei, Furu},
  journal={arXiv preprint arXiv:2106.08254},
  year={2021}
}

@article{liu2020understanding,
  title={Understanding the difficulty of training transformers},
  author={Liu, Liyuan and Liu, Xiaodong and Gao, Jianfeng and Chen, Weizhu and Han, Jiawei},
  journal={arXiv preprint arXiv:2004.08249},
  year={2020}
}

@inproceedings{xiong2020layer,
  title={On layer normalization in the transformer architecture},
  author={Xiong, Ruibin and Yang, Yunchang and He, Di and Zheng, Kai and Zheng, Shuxin and Xing, Chen and Zhang, Huishuai and Lan, Yanyan and Wang, Liwei and Liu, Tieyan},
  booktitle={International conference on machine learning},
  pages={10524--10533},
  year={2020},
  organization={PMLR}
}

@inproceedings{glorot2010understanding,
  title={Understanding the difficulty of training deep feedforward neural networks},
  author={Glorot, Xavier and Bengio, Yoshua},
  booktitle={Proceedings of the thirteenth international conference on artificial intelligence and statistics},
  pages={249--256},
  year={2010},
  organization={JMLR Workshop and Conference Proceedings}
}
}

\clearpage
\setcounter{page}{1}
\maketitlesupplementary

\section{Backward Pass Derivation}
\label{app:warp-disc-bwd-math}

Using the warping definition of the forward pass,

\begin{equation}
    y_{n\mathbf{v}} = \sum_\mathbf{u}x_{n\mathbf{u}}\prod_{k}\varphi\left(\mathbf{v}_k-\mathbf{u}_k+\varepsilon_{k\mathbf{v}}\right)
\end{equation}

The gradient with respect to $x$ can be determined as follows,

\begin{equation}
\begin{split}
    \nabla_{x} \mathcal{L}_{iw}
    &= \partial_{x_{iw}} \mathcal{L}
     = \sum_{n,\mathbf{v}} \nabla_y\mathcal{L}_{n\mathbf{v}}\partial_{x_{i\mathbf{w}}} y_{n\mathbf{v}}\\
    &= \sum_{n,\mathbf{v}} \nabla_y\mathcal{L}_{n\mathbf{v}}\sum_\mathbf{u}\partial_{x_{i\mathbf{w}}}x_{n\mathbf{u}}\prod_{k}\varphi\left(\mathbf{v}_k-\mathbf{u}_k+\varepsilon_{k\mathbf{v}}\right)\\
    &= \sum_{n,\mathbf{v}} \nabla_y\mathcal{L}_{n\mathbf{v}}\sum_\mathbf{u}\delta_{in}\delta_{\mathbf{wu}}\prod_{k}\varphi\left(\mathbf{v}_k-\mathbf{u}_k+\varepsilon_{k\mathbf{v}}\right)\\
    &= \sum_{\mathbf{v}} \nabla_y\mathcal{L}_{i\mathbf{v}}\prod_{k}\varphi\left(\mathbf{v}_k-\mathbf{w}_k+\varepsilon_{k\mathbf{v}}\right)\\
\end{split}
\end{equation}

And the gradient with respect to $\varepsilon$ can be determined as follows,

\begin{equation}
\begin{split}
    \nabla_\varepsilon \mathcal{L}_{i\mathbf{w}}
    &= \partial_{\varepsilon_{i\mathbf{w}}} \mathcal{L}
     = \sum_{n,\mathbf{v}} \nabla_y\mathcal{L}_{n\mathbf{v}}\partial_{\varepsilon_{i\mathbf{w}}} y_{n\mathbf{v}}\\
    &= \sum_{n,\mathbf{v}} \nabla_y\mathcal{L}_{n\mathbf{v}}\sum_\mathbf{u}x_{n\mathbf{u}} \partial_{\varepsilon_{i\mathbf{w}}}\prod_{k}\varphi\left(\mathbf{v}_k-\mathbf{u}_k+\varepsilon_{k\mathbf{v}}\right)\\
    &= \sum_{n,\mathbf{v}} \nabla_y\mathcal{L}_{n\mathbf{v}}\sum_\mathbf{u}x_{n\mathbf{u}} \partial_{\varepsilon_{i\mathbf{w}}}\varphi\left(\mathbf{v}_i-\mathbf{u}_i+\varepsilon_{i\mathbf{v}}\right) \prod_{k\neq i}\varphi\left(\mathbf{v}_k-\mathbf{u}_k+\varepsilon_{k\mathbf{v}}\right)\\
    &= \sum_{n,\mathbf{v}} \nabla_y\mathcal{L}_{n\mathbf{v}}\sum_\mathbf{u}x_{n\mathbf{u}} \varphi^\prime\left(\mathbf{v}_i-\mathbf{u}_i+\varepsilon_{i\mathbf{v}}\right) \delta_{\mathbf{vw}} \prod_{k\neq i}\varphi\left(\mathbf{v}_k-\mathbf{u}_k+\varepsilon_{k\mathbf{v}}\right)\\
    &= \sum_{n} \nabla_y\mathcal{L}_{n\mathbf{w}}\sum_\mathbf{u}x_{n\mathbf{u}} \varphi^\prime\left(\mathbf{w}_i-\mathbf{u}_i+\varepsilon_{i\mathbf{w}}\right) \prod_{k\neq i}\varphi\left(\mathbf{w}_k-\mathbf{u}_k+\varepsilon_{k\mathbf{w}}\right)\\
\end{split}
\end{equation}

\section{IID Analysis}
\label{sec:app.iid-analysis}

For this analysis, we assume 3 properties on the design of $\varphi(t)$.  The first is that in the case where $\varepsilon = \mathbf{0}$, we want the warping operation to be an identity function.  This implies that for every non-zero integer $n$, $\varphi(n)=0$, and $\varphi(0)=1$.  Secondly, to prevent a quadratic computation cost, we enforce that $\varphi(t)=0$ for all $|t|>1$.  Lastly, there would be no reason to directionally specialize the interpolation, so we enforce $\varphi(t) = \varphi(-t)$.

Using these properties, the iid analysis follows.

\begin{equation}
\begin{split}
    \mu_y
    &=\Ex{y_{nv}} = \Ex{\sum_u x_{nu}\prod_k\varphi(v_k-u_k+\varepsilon_{kv})}\\
    &=\sum_u\Ex{x_{nu}}\Ex{\prod_k\varphi(v_k-u_k+\varepsilon_{kv})}\\
    &=\mu_x\sum_u\prod_k\Ex{\varphi(v_k-u_k+\varepsilon_{kv})}\\
    &=\mu_x\prod_k\sum_u\Ex{\varphi(v_k-u+\varepsilon_{kv})}\\
    &=\mu_x\prod_k\Ex{\sum_u\varphi(v_k-u+\varepsilon_{kv})}\\
    &=\mu_x\Phi_{1,1}^d
\end{split}
\end{equation}

\begin{equation}
\begin{split}
    \mu_y^2 + \sigma_y^2
    &= \Ex{y_{nv}^2} = \Ex{\left(\sum_u x_{nu}\prod_k\varphi(v_k-u_k+\varepsilon_{kv})\right)^2}\\
    &= \Ex{\sum_{rs}x_{nr}x_{ns}\prod_k\varphi(v_k-r_k+\varepsilon_{kv})\varphi(v_k-s_k+\varepsilon_{kv})}\\
    &= \Ex{\sum_{r\neq s}x_{nr}x_{ns}\prod_k\varphi(v_k-r_k+\varepsilon_{kv})\varphi(v_k-s_k+\varepsilon_{kv})+\sum_u x_{nu}^2\prod_k\varphi^2(v_k-u_k+\varepsilon_{kv})}\\
    &= \sum_{r\neq s}\mu_x^2\Ex{\prod_k\varphi(v_k-r_k+\varepsilon_{kv})\varphi(v_k-s_k+\varepsilon_{kv})}+\sum_u(\sigma_x^2+\mu_x^2)\Ex{\prod_k\varphi^2(v_k-u_k+\varepsilon_{kv})}\\
    &= \mu_x^2\Ex{\sum_{r\neq s}\prod_k\varphi(v_k-r_k+\varepsilon_{kv})\varphi(v_k-s_k+\varepsilon_{kv})}+(\sigma_x^2+\mu_x^2)\Ex{\sum_u\prod_k\varphi^2(v_k-u_k+\varepsilon_{kv})}\\
    &= \mu_x^2\Ex{\sum_{rs}\prod_k\varphi(v_k-r_k+\varepsilon_{kv})\varphi(v_k-s_k+\varepsilon_{kv})-\sum_u\prod_k\varphi^2(v_k-u_k+\varepsilon_{kv})}\\
    &\hspace{50pt}+(\sigma_x^2+\mu_x^2)\prod_k\Ex{\sum_u\varphi^2(v_k-u+\varepsilon_{kv})}\\
    &= \mu_x^2\Ex{\left(\sum_u\prod_k\varphi(v_k-u_k+\varepsilon_{kv})\right)^2-\sum_u\prod_k\varphi^2(v_k-u_k+\varepsilon_{kv})} + (\sigma_x^2+\mu_x^2)\Phi_{2,1}^d\\
    &= \mu_x^2\Ex{\left(\prod_k\sum_u\varphi(v_k-u+\varepsilon_{kv})\right)^2-\prod_k\sum_u\varphi^2(v_k-u+\varepsilon_{kv})} + (\sigma_x^2+\mu_x^2)\Phi_{2,1}^d\\
    &= \mu_x^2\prod_k\Ex{\left(\sum_u\varphi(v_k-u+\varepsilon_{kv})\right)^2}-\mu_x^2\prod_k\Ex{\sum_u\varphi^2(v_k-u+\varepsilon_{kv})} + (\sigma_x^2+\mu_x^2)\Phi_{2,1}^d\\
    &= \mu_x^2\Phi_{1,2}^d-\mu_x^2\Phi_{2,1}^d+(\sigma_x^2+\mu_x^2)\Phi_{2,1}^d\\
    &= \mu_x^2\Phi_{1,2}^d+\sigma_x^2\Phi_{2,1}^d
\end{split}
\end{equation}

\begin{equation}
\begin{split}
    \mu_{\partial x}
    &=\Ex{\nabla_{x} \mathcal{L}_{iw}} = \Ex{\sum_{v} \nabla_y\mathcal{L}_{iv}\prod_{k}\varphi\left(u_k-v_k+\varepsilon_{ku}\right)}\\
    &=\sum_{v} \Ex{\nabla_y\mathcal{L}_{iv}}\Ex{\prod_{k}\varphi\left(u_k-v_k+\varepsilon_{ku}\right)}\\
    &=\mu_{\partial y}\sum_{v}\prod_{k}\Ex{\varphi\left(u_k-v_k+\varepsilon_{ku}\right)}\\
    &=\mu_{\partial y}\prod_{k}\sum_{v}\Ex{\varphi\left(u_k-v+\varepsilon_{ku}\right)}\\
    &=\mu_{\partial y}\prod_{k}\Ex{\sum_{v}\varphi\left(u_k-v+\varepsilon_{ku}\right)}\\
    &=\mu_{\partial y}\Phi_{1,1}^d\\
\end{split}
\end{equation}

\begin{equation}
\begin{split}
    \mu_{\partial x}^2+\sigma_{\partial x}^2
    &=\Ex{\nabla_x\mathcal{L}_{iw}^2} = \Ex{\left(\sum_v\nabla_y\mathcal{L}_{iv}\prod_k\varphi(u_k-v_k+\varepsilon_{ku}\right)^2}\\
    &=\Ex{\sum_{r,s}\nabla_y\mathcal{L}_{ir}\nabla_y\mathcal{L}_{is}\prod_k\varphi(r_k-w_k+\varepsilon_{kr})\varphi(s_k-w_k+\varepsilon_{ks})}\\
\end{split}
\end{equation}

\section{Discretization Effects}
\label{app:discrete-effects}

We enforce three properties on the design of $\varphi(t)$.  The first is that in the case where $\varepsilon = \mathbf{0}$, we want the warping operation to be an identity function.  This implies that for every non-zero integer $n$, $\varphi(n)=0$, and $\varphi(0)=1$.  Secondly, to prevent a quadratic computation cost, we enforce that $\varphi(t)=0$ for all $|t|>1$.  Lastly, there would be no reason to directionally specialize the interpolation, so we enforce $\varphi(t) = \varphi(-t)$.

Two cases emerge from these equations.  If mean is preserved, $\mu_y=\mu_x$ results in a constraint of $\varphi(t) + \varphi(t+1) = 1$.  If a mean of 0 is assumed, and variance is preserved, $\sigma_y^2=\sigma_x^2$ results in $\varphi(t)^2 + \varphi(t+1)^2 = 1$.  Unfortunately, it is only possible to satisfy both of these constraints if $\varphi(t)$ is the rectangle function.  This case corresponds to nearest-neighbor interpolation between pixels, and also results in $\nabla_\varepsilon\mathcal{L} = \mathbf{0}$.  For this reason, we design $\varphi(t)$ to only satisfy one constraint in practice.  Also note, that for any interpolation kernel $\varphi(t)$ that is mean-preserving, the square root of the kernel will be variance preserving.

To implement bi-linear interpolation, one can take $\varphi(t)=\text{tri}(t)$.  This choice results in the following statistical properties,

\begin{equation}
\begin{gathered}
    \Phi_{1,1}=\Phi_{1,2}=1 \quad
    \Phi_{1,1}^\prime=\Phi_{1,2}^\prime = 0 \quad
    \Phi_{2,1}^\prime = 2 \\
    \Phi_{2,1} = \frac{2}{3}+2\sum_{n=1}^\infty\frac{e^{-2\pi^2n^2\sigma_\varepsilon^2}}{\pi^2n^2}
\end{gathered}
\end{equation}

As an example of a variance preserving interpolation kernel, we use $\phi(t)=\text{cos}(\pi t/2)$.  This choice results in the following statistical characterization.

\begin{equation}
    \Phi_{2,1} = 1 \quad
    \Phi_{2,1}^\prime = \frac{\pi^2}{4}
\end{equation}

If used directly, these variance characteristics ensure that any network structure that uses warping blocks will lead to exploding gradients.  We believe this is the reason why deformable convolutions needed to use both a pre-trained network and decrease the learning rate for the sampling parameters.

\section{Random Field Derivation}
\label{app:random-field}

\begin{equation}
    y(t) = x(t+\varepsilon(t))
\end{equation}

\begin{equation}
    \begin{split}
    R_y(t)
    &= \Ex[y]{y(\tau)y(t+\tau)} \\
    &= \Ex[x,\varepsilon]{x(\tau+\varepsilon(\tau))x(t+\tau+\varepsilon(t+\tau))}\\
    &= \Ex[u,v\sim p(u,v|t)]{\Ex[x]{x(\tau+u)x(t+\tau+v)}}\\
    &= \Ex[u,v]{R_x(t-v+u)}
    \end{split}
\end{equation}

\begin{equation}
    \begin{split}
    p(u,v|t)
    &= \prod_k p(u_k,v_k|t)\\
    &= \prod_k\int_0^1\gamma_t(\rho)\mathcal{N}\left(u_k,v_k\left|0,\begin{bmatrix}
            \sigma_\varepsilon^2 & \rho\sigma_\varepsilon^2 \\ \rho\sigma_\varepsilon^2 & \sigma_\varepsilon^2
        \end{bmatrix}\right.\right)d\rho\\
    &= \int_{[0,1]^N}\prod_k\gamma_t(\rho_k)\mathcal{N}\left(u_k,v_k\left|0,\begin{bmatrix}
            \sigma_\varepsilon^2 & \rho_k\sigma_\varepsilon^2 \\ \rho_k\sigma_\varepsilon^2 & \sigma_\varepsilon^2
        \end{bmatrix}\right.\right)d\rho
    \end{split}
\end{equation}

Where $\forall t$, $\langle\gamma_t(p)\rangle=1$.  This was chosen so that the marginal distribution is $\forall\gamma_t(\rho)$, $p(u)=\prod_k\mathcal{N}(u_k|0,\sigma_\varepsilon^2)$.

\begin{equation}
    \begin{split}
    \mathcal{N}\left(u_k,v_k\left|0,
    \begin{bmatrix}
        \sigma_\varepsilon^2 & \rho_k\sigma_\varepsilon^2 \\ \rho_k\sigma_\varepsilon^2 & \sigma_\varepsilon^2
    \end{bmatrix}\right.\right)
    &= \frac{1}{2\pi\sigma_\varepsilon^2\sqrt{1-\rho_k^2}}\exp{\left(
        -\frac{u_k^2 - 2\rho_k u_k v_k + v_k^2}{2\sigma_\varepsilon^2\left(1-\rho_k^2\right)}\right)}\\
    &= \frac{1}{2\pi\sigma_\varepsilon^2\sqrt{1-\rho_k^2}}\exp{\left(
        -\frac{\left(u_k-v_k\right)^2}{8\sigma_\varepsilon^2\left(1-\rho_k\right)}
        -\frac{\left(u_k+v_k\right)^2}{8\sigma_\varepsilon^2\left(1+\rho_k\right)}\right)}\\
    &=  \frac{1}{\sigma_\varepsilon\sqrt{1-\rho_k}\sqrt{2\pi}}\exp{\left(
        -\frac{\left(u_k-v_k\right)^2}{8\sigma_\varepsilon^2\left(1-\rho_k\right)}\right)}
        \frac{1}{\sigma_\varepsilon\sqrt{1+\rho_k}\sqrt{2\pi}}\exp{\left(
        -\frac{\left(u_k+v_k\right)^2}{8\sigma_\varepsilon^2\left(1+\rho_k\right)}\right)}\\
    &=  2\mathcal{N}\left(u_k+v_k|0,2\sigma_\varepsilon^2\left(1+\rho_k\right)\right)
        \mathcal{N}\left(u_k-v_k|0,2\sigma_\varepsilon^2\left(1-\rho_k\right)\right)
    \end{split}
\end{equation}

Using this substitution, we can directly write the autocorrelation $R_y(t)$ in terms of the shaping function $\gamma_t(\rho)$.

\begin{equation}
    \begin{split}
    R_y(t)
    &= \int_{\mathbb{R}^{2N}}R_x(t-v+u)p(u,v|t)dudv \\
    &= \int_{\mathbb{R}^{2N}}R_x(t-v+u)\int_{[0,1]^N}
        \prod_k\gamma_t(\rho_k)
        \mathcal{N}\left(u_k,v_k\left|0,
        \begin{bmatrix}
            \sigma_\varepsilon^2 & \rho_k\sigma_\varepsilon^2 \\ \rho_k\sigma_\varepsilon^2 & \sigma_\varepsilon^2
        \end{bmatrix}
        \right.\right)
    d\rho dudv\\
    &= \int_{[0,1]^N}\left(\prod_k\gamma_t(\rho_k)\right)\int_{\mathbb{R}^{2N}}R_x(t-v+u)
        \prod_k\mathcal{N}\left(u_k,v_k\left|0,
        \begin{bmatrix}
            \sigma_\varepsilon^2 & \rho_k\sigma_\varepsilon^2 \\ \rho_k\sigma_\varepsilon^2 & \sigma_\varepsilon^2
        \end{bmatrix}
        \right.\right)
    dudvd\rho\\
    &= \int_{[0,1]^N}\left(\prod_k\gamma_t(\rho_k)\right)\int_{\mathbb{R}^{2N}}R_x(t-v+u)\prod_k2
        \mathcal{N}\left(u_k+v_k|0,2\sigma_\varepsilon^2\left(1+\rho_k\right)\right)
        \mathcal{N}\left(u_k-v_k|0,2\sigma_\varepsilon^2\left(1-\rho_k\right)\right)
    dudvd\rho\\
    &= \int_{[0,1]^N}\left(\prod_k\gamma_t(\rho_k)\right)\int_{\mathbb{R}^{2N}}R_x(t-u)
        \frac{1}{2^N}\prod_k2
            \mathcal{N}\left(u_k|0,2\sigma_\varepsilon^2\left(1-\rho_k\right)\right)
            \mathcal{N}\left(v_k|0,2\sigma_\varepsilon^2\left(1+\rho_k\right)\right)
    dudvd\rho\\
    &= \int_{[0,1]^N}\left(\prod_k\gamma_t(\rho_k)\right)\int_{\mathbb{R}^{N}}R_x(t-u)
        \prod_k
            \mathcal{N}\left(\left.u_k\right|0,2\sigma_\varepsilon^2\left(1-\rho_k\right)\right)
            \int_{-\infty}^\infty\mathcal{N}\left(v|0,2\sigma_\varepsilon^2\left(1+\rho_k\right)\right)dv
    dud\rho\\
    &= \int_{[0,1]^N}\left(\prod_k\gamma_t(\rho_k)\right)\int_{\mathbb{R}^{N}}R_x(t-u)
        \prod_k\mathcal{N}\left(\left.u_k\right|0,2\sigma_\varepsilon^2\left(1-\rho\right)\right)
    dud\rho\\
    &= \int_{\mathbb{R}^{N}}R_x(t-u)
        \prod_k\int_0^1\gamma_t(\rho)\mathcal{N}\left(\left.u_k\right|0,2\sigma_\varepsilon^2\left(1-\rho\right)\right)
    d\rho du
    \end{split}
\end{equation}

\begin{equation}
    \begin{split}
    S_y(\xi)
    &= \mathcal{F}\{R_y(t)\} = \int_{\mathbb{R}^N}R_y(t)e^{-2\pi i \xi\cdot t}dt\\
    &= \int_{\mathbb{R}^N}e^{-2\pi i \xi\cdot t}\int_{\mathbb{R}^N}R_x(t-u)
        \prod_k\int_0^1\gamma_t(\rho)\mathcal{N}\left(\left.u_k\right|0,2\sigma_\varepsilon^2\left(1-\rho\right)\right)
        d\rho dudt\\
    &= \int_{[0,1]^N}\int_{\mathbb{R}^{2N}}e^{-2\pi i \xi\cdot t}R_x(t-u)
        \left(\prod_k\gamma_t(\rho_k)\right)\left(\prod_k\mathcal{N}(u_k|0,2\sigma_\varepsilon^2(1-\rho_k))\right)
        dtdud\rho\\
    &= \int_{[0,1]^N}\int_{\mathbb{R}^{2N}}
        \left(\prod_k\gamma_{u+v}(\rho_k)\right)
        e^{-2\pi i \xi\cdot v}R_x(v)
        e^{-2\pi i \xi\cdot u}\left(\prod_k\mathcal{N}(u_k|0,2\sigma_\varepsilon^2(1-\rho_k))\right)
        dvdud\rho
    \end{split}
\end{equation}

Using a Fourier Series representation of $\gamma_t(\rho)$, the $u+v$ term in the PSD expression can be separated into a product of functions of $v$ and functions of $u$.

\begin{equation}
    \begin{split}
    \prod_k\gamma_t(\rho_k)
    &= \prod_k\sum_{j\in\mathbb{Z}}\sum_{\omega\in\mathbb{Z}^N}c_{j\omega}e^{2\pi ij\rho_k}e^{2\pi i\omega\cdot t}\\
    &= \sum_{j\in\mathbb{Z}^N}\sum_{\omega\in\mathbb{Z}^{N^2}}\prod_kc_{j_k\omega_k}e^{2\pi ij_k\rho_k}e^{2\pi i\omega_k\cdot t}\\
    &= \sum_{j\in\mathbb{Z}^N}\sum_{\omega\in\mathbb{Z}^{N^2}}\hat{c_{j\omega}}e^{2\pi i\hat{j}\cdot\rho}e^{2\pi i\hat{\omega}\cdot t}\\
    & \hat{c_{j\omega}} = \prod_kc_{j_k\omega_k} \hspace{10pt} \hat{j} = \sum_kj_k \hspace{10pt} \hat{\omega} = \sum_k\omega_k
    \end{split}
\end{equation}

Using this representation, the integrals over $u$ and $v$ can be factored.

\begin{equation}
    \begin{split}
    S_y(\xi)
    &= \sum_{j\in\mathbb{Z^N}}\sum_{\omega\in\mathbb{Z}^{N^2}}\hat{c_{j\omega}}\int_{[0,1]^N}\int_{\mathbb{R}^{2N}}
        e^{2\pi i\hat{j}\cdot\rho}e^{2\pi i\hat{\omega}\cdot (u+v)}
        e^{-2\pi i \xi\cdot v}R_x(v)
        e^{-2\pi i \xi\cdot u}\left(\prod_k\mathcal{N}(u_k|0,2\sigma_\varepsilon^2(1-\rho_k))\right)
        dvdud\rho\\
    &= \sum_{j\in\mathbb{Z^N}}\sum_{\omega\in\mathbb{Z}^{N^2}}\hat{c_{j\omega}}\int_{[0,1]^N}e^{2\pi i\hat{j}\cdot\rho}
        \int_{\mathbb{R}^N}e^{-2\pi i (\xi-\hat{\omega})\cdot v}R_x(v)dv
        \int_{\mathbb{R}^N}e^{-2\pi i (\xi-\hat{\omega})\cdot u}\left(\prod_k\mathcal{N}(u_k|0,2\sigma_\varepsilon^2(1-\rho_k))\right)du
        d\rho\\
    &= \sum_{\omega\in\mathbb{Z}^{N^2}}S_x(\xi-\hat{\omega})\sum_{j\in\mathbb{Z}^N}\hat{c_{j\omega}}\int_{[0,1]^N}e^{2\pi i\hat{j}\cdot\rho}
        \int_{\mathbb{R}^N}e^{-2\pi i (\xi-\hat{\omega})\cdot u}\left(\prod_k\mathcal{N}(u_k|0,2\sigma_\varepsilon^2(1-\rho_k))\right)du
        d\rho\\
    &= \sum_{\omega\in\mathbb{Z}^{N^2}}S_x(\xi-\hat{\omega})\sum_{j\in\mathbb{Z}^N}\hat{c_{j\omega}}\int_{[0,1]^N}e^{2\pi i\hat{j}\cdot\rho}
        \left(\prod_k\int_{\mathbb{R}}e^{-2\pi i (\xi_k-\hat{\omega}_k)u_k}\mathcal{N}(u_k|0,2\sigma_\varepsilon^2(1-\rho_k))du_k\right)
        d\rho\\
    &= \sum_{\omega\in\mathbb{Z}^{N^2}}S_x(\xi-\hat{\omega})\sum_{j\in\mathbb{Z}^N}\hat{c_{j\omega}}\int_{[0,1]^N}
        \left(\prod_ke^{2\pi ij_k\rho_k}\right)
        \left(\prod_ke^{-4\pi^2\sigma_\varepsilon^2(\xi_k-\hat{\omega}_k)^2(1-\rho_k)}\right)
        d\rho\\
    &= \sum_{\omega\in\mathbb{Z}^{N^2}}S_x(\xi-\hat{\omega})\sum_{j\in\mathbb{Z}^N}\hat{c_{j\omega}}
        \prod_k\int_0^1e^{2\pi ij_k\rho_k}e^{-4\pi^2\sigma_\varepsilon^2(\xi_k-\hat{\omega}_k)^2(1-\rho_k)}
        d\rho\\
    &= \sum_{\omega\in\mathbb{Z}^{N^2}}S_x(\xi-\hat{\omega})\sum_{j\in\mathbb{Z}^N}\hat{c_{j\omega}}
        \prod_ke^{-4\pi^2\sigma_\varepsilon^2(\xi_k-\hat{\omega}_k)^2}
        \int_0^1e^{(2\pi ij_k+4\pi^2\sigma_\varepsilon^2(\xi_k-\hat{\omega}_k)^2)\rho_k}
        d\rho\\
    &= \sum_{\omega\in\mathbb{Z}^{N^2}}S_x(\xi-\hat{\omega})\sum_{j\in\mathbb{Z}^N}\hat{c_{j\omega}}
        \prod_ke^{-4\pi^2\sigma_\varepsilon^2(\xi_k-\hat{\omega}_k)^2}
        \frac{e^{4\pi^2\sigma_\varepsilon^2(\xi_k-\hat{\omega}_k)^2}-1}
        {2\pi ij_k+4\pi^2\sigma_\varepsilon^2(\xi_k-\hat{\omega}_k)^2}\\
    &= \sum_{\omega\in\mathbb{Z}^{N^2}}S_x(\xi-\hat{\omega})\sum_{j\in\mathbb{Z}^N}
        \left(\prod_kc_{j_k\omega_k}\right)
        \left(\prod_k\frac{1-e^{-4\pi^2\sigma_\varepsilon^2(\xi_k-\hat{\omega}_k)^2}}
        {2\pi ij_k+4\pi^2\sigma_\varepsilon^2(\xi_k-\hat{\omega}_k)^2}\right)\\
    &= \sum_{\omega\in\mathbb{Z}^{N^2}}S_x(\xi-\hat{\omega})
        \prod_k\sum_{j\in\mathbb{Z}}c_{j_k\omega_k}
        \frac{1-e^{-4\pi^2\sigma_\varepsilon^2(\xi_k-\hat{\omega}_k)^2}}
        {2\pi ij_k+4\pi^2\sigma_\varepsilon^2(\xi_k-\hat{\omega}_k)^2}\\
    &= \sum_{\omega\in\mathbb{Z}^{N^2}}S_x(\xi-\hat{\omega})
        \prod_k\left(1-e^{-4\pi^2\sigma_\varepsilon^2(\xi_k-\hat{\omega}_k)^2}\right)
        \sum_{j\in\mathbb{Z}}\frac{c_{j\omega_k}}
        {2\pi ij+4\pi^2\sigma_\varepsilon^2(\xi_k-\hat{\omega}_k)^2}\\
    \end{split}
\end{equation}

Note that in the above analysis, Fubini's theorem is violated for certain choices of $\gamma_t(\rho)$, such as $\delta(\rho-f(t))$.  Let $z_j=2\pi ij$, $r_k=4\pi^2\sigma_\varepsilon^2(\xi_k-\hat{\omega_k})^2$.

\begin{equation}
    \begin{split}
    S_y(\xi)
    &= \sum_{\omega\in\mathbb{Z}^{N^2}}S_x(\xi-\hat{\omega})
        \prod_k\left(1-e^{-r_k}\right)\sum_{j\in\mathbb{Z}}\frac{c_{j\omega_k}}{z_j+r_k}\\
    &= \sum_{\omega\in\mathbb{Z}^{N^2}}S_x(\xi-\hat{\omega})
        \prod_k\left(1-e^{-r_k}\right)\sum_{j\in\mathbb{Z}}\sum_{m=1}^\infty c_{j\omega_k}(-r_k)^{m-1}z_j^{-m}\\
    &= \sum_{\omega\in\mathbb{Z}^{N^2}}S_x(\xi-\hat{\omega})
        \prod_k\left(1-e^{-r_k}\right)\sum_{m=1}^\infty(-r_k)^{m-1}\sum_{j\in\mathbb{Z}}c_{j\omega_k}z_j^{-m}\\
    \end{split}
\end{equation}

Let $a_{m\omega}$ be the $m^{th}$ moment of the $\omega^{th}$ Fourier coefficient over $t$ of $\gamma_t(\rho)$.

\begin{equation}
    \begin{split}
    a_{mn}
    &= \int_{[0,1]^N}\int_0^1\rho^m\gamma_t(\rho)d\rho e^{-2\pi in\cdot t}dt\\
    &= \sum_{j\in\mathbb{Z}}\sum_{\omega\in\mathbb{Z}^N}c_{j\omega}\int_{[0,1]^N}\int_0^1\rho^m
        e^{2\pi ij\rho}e^{2\pi i\omega\cdot t}d\rho e^{-2\pi in\cdot t}dt\\
    &= \sum_{j\in\mathbb{Z}}\sum_{\omega\in\mathbb{Z}^N}c_{j\omega}
        \int_0^1\rho^me^{2\pi ij\rho}d\rho\int_{[0,1]^N}e^{2\pi i(\omega-n)\cdot t}dt\\
    &= \sum_{j\in\mathbb{Z}}\sum_{\omega\in\mathbb{Z}^N}c_{j\omega}
        \int_0^1\rho^me^{z_j\rho}d\rho\delta_{\omega n}
        = \sum_{j\in\mathbb{Z}}c_{jn}\int_0^1\rho^me^{z_j\rho}d\rho\\
    &= \sum_{j\in\mathbb{Z}}c_{jn}I_j(m) \hspace{10pt} \text{where} \hspace{10pt} I_j(m) = \int_0^1\rho^me^{z_j\rho}d\rho
    \end{split}
\end{equation}

One can apply integration by parts to the $I_j(m)$ definition to obtain the following recurrence relation.

\begin{equation}
    \begin{split}
    I_j(m) = \frac{1-mI_j(m-1)}{z_j} \hspace{10pt} I_j(1)=\frac{1}{z_j}
    \end{split}
\end{equation}

Solving this relation results in the following closed form expression for $I_j(m)$.  It is easily verified that the recurrence relation follows from closed form expression.

\begin{equation}
    I_j(m) = -\sum_{n=1}^m\frac{m!}{n!}(-z_j)^{n-m-1}
\end{equation}

This expression can be inverted to provide an expression for $z_j^{-m}$, where $B_n^-$ represents the $n^{th}$ negative Bernoulli number.

\begin{equation}
    z_j^{-m}=(-1)^{m-1}\sum_{n=0}^{m-1}\frac{I_j(m-n)B_n^-}{(m-n)!n!}
\end{equation}

This equation can be verified using induction.

\begin{equation}
    \begin{split}
    z_j^{-m-1}
    &= (-1)^m\sum_{n=0}^m\frac{I_j(m-n+1)B_n^-}{(m-n+1)!n!}\\
    &= \sum_{n=0}^m(-1)^{m-n}\frac{I_j(m-n+1)B_n^+}{(m-n+1)!n!}\\
    &= \frac{B_m^+}{z_jm!}+\sum_{n=0}^{m-1}(-1)^{m-n}\frac{I_j(m-n+1)B_n^+}{(m-n+1)!n!}\\
    &= \frac{1}{z_j}\left(\frac{B_m^+}{m!}+\sum_{n=0}^{m-1}(-1)^{m-n}\frac{\left(1-(m-n+1)I_j(m-n)\right)B_n^+}{(m-n+1)!n!}\right)\\
    &= \frac{1}{z_j}\left(\frac{B_m^+}{m!}+\sum_{n=0}^{m-1}\frac{(-1)^{m-n}B_n^+}{(m-n+1)!n!}-\sum_{n=0}^{m-1}(-1)^{m-n}\frac{I_j(m-n)B_n^+}{(m-n)!n!}\right)\\
    &= \frac{1}{z_j}\left(\frac{B_m^+}{m!}+(-1)^m\sum_{n=0}^{m-1}\frac{B_n^-}{(m-n+1)!n!}+z_j^{-m}\right)\\
    &= \frac{1}{z_j}\left(\frac{B_m^+}{m!}-(-1)^m\frac{B_m^-}{m!}+z_j^{-m}\right)\\
    &= \frac{1}{z_j}\left(\frac{B_m^+}{m!}-\frac{B_m^+}{m!}+z_j^{-m}\right)=z_j^{-m-1}\\
    \end{split}
\end{equation}

\begin{equation}
    \begin{split}
    S_y(\xi)
    &= \sum_{\omega\in\mathbb{Z}^{N^2}}S_x(\xi-\hat{\omega})
        \prod_k\left(1-e^{-r_k}\right)\sum_{m=1}^\infty(-r_k)^{m-1}\sum_{j\in\mathbb{Z}}c_{j\omega_k}z_j^{-m}\\
    &= \sum_{\omega\in\mathbb{Z}^{N^2}}S_x(\xi-\hat{\omega})
        \prod_k\left(1-e^{-r_k}\right)\sum_{m=1}^\infty(-r_k)^{m-1}\sum_{j\in\mathbb{Z}}c_{j\omega_k}(-1)^{m-1}\sum_{n=0}^{m-1}\frac{I_j(m-n)B_n^-}{(m-n)!n!}\\
    &= \sum_{\omega\in\mathbb{Z}^{N^2}}S_x(\xi-\hat{\omega})
        \prod_k\left(1-e^{-r_k}\right)\sum_{m=1}^\infty\sum_{n=0}^{m-1}\frac{r_k^{m-1}B_n^-}{(m-n)!n!}\sum_{j\in\mathbb{Z}}c_{j\omega_k}I_j(m-n)\\
    &= \sum_{\omega\in\mathbb{Z}^{N^2}}S_x(\xi-\hat{\omega})
        \prod_k\left(1-e^{-r_k}\right)\sum_{m=1}^\infty\sum_{n=0}^{m-1}\frac{r_k^{m-1}B_n^-a_{m-n,\omega_k}}{(m-n)!n!}\\
    &= \sum_{\omega\in\mathbb{Z}^{N^2}}S_x(\xi-\hat{\omega})
        \prod_k\left(1-e^{-r_k}\right)\sum_{m=1}^\infty\sum_{n=0}^\infty\frac{r_k^{m+n-1}B_n^-a_{m\omega_k}}{m!n!}\\
    &= \sum_{\omega\in\mathbb{Z}^{N^2}}S_x(\xi-\hat{\omega})
        \prod_k\left(1-e^{-r_k}\right)\left(\sum_{m=1}^\infty\frac{r_k^{m}a_{m\omega_k}}{m!}\right)\left(\sum_{n=0}^\infty\frac{B_n^-r_k^{n-1}}{n!}\right)\\
    &= \sum_{\omega\in\mathbb{Z}^{N^2}}S_x(\xi-\hat{\omega})
        \prod_k\frac{1-e^{-r_k}}{e^{r_k}-1}\sum_{m=1}^\infty\frac{r_k^{m}a_{m\omega_k}}{m!}\\
    &= \sum_{\omega\in\mathbb{Z}^{N^2}}S_x(\xi-\hat{\omega})
        \prod_ke^{-r_k}\sum_{m=1}^\infty\frac{r_k^{m}a_{m\omega_k}}{m!}\\
    \end{split}
\end{equation}

If we take $\varepsilon(t)$ to be a gaussian random process, then $\gamma_t(\rho)=\delta(\rho-f(t))$ to ensure that the joint distribution over any two samples of $\varepsilon$ always collapses to a joint gaussian, where $f(t)$ controls the correlation at time delay $t$.

\begin{equation}
    \begin{split}
    a_{mn}
    &= \int_{[0,1]^N}\int_0^1\rho^m\gamma_t(\rho)d\rho e^{-2\pi in\cdot t}dt\\
    &= \int_{[0,1]^N}\int_0^1\rho^m\delta(\rho-f(t))d\rho e^{-2\pi in\cdot t}dt\\
    &= \int_{[0,1]^N}f(t)^m e^{-2\pi in\cdot t}dt\\
    \end{split}
\end{equation}

Substituting this equation into the power spectral density equation:

\begin{equation}
    \begin{split}
    S_y(\xi)
    &= \sum_{\omega\in\mathbb{Z}^{N^2}}S_x(\xi-\hat{\omega})
        \prod_ke^{-r_k}\sum_{m=1}^\infty\frac{r_k^{m}a_{m\omega_k}}{m!}\\
    &= \sum_{\omega\in\mathbb{Z}^{N^2}}S_x(\xi-\hat{\omega})
        \prod_ke^{-r_k}\sum_{m=1}^\infty\frac{r_k^{m}}{m!}\int_{[0,1]^N}f(t)^m e^{-2\pi i\omega_k\cdot t}dt\\
    &= \sum_{\omega\in\mathbb{Z}^{N^2}}S_x(\xi-\hat{\omega})
        \prod_ke^{-r_k}\int_{[0,1]^N}e^{-2\pi i\omega_k\cdot t}\sum_{m=1}^\infty\frac{(r_kf(t))^{m}}{m!}dt\\
    &\approx \sum_{\omega\in\mathbb{Z}^{N^2}}S_x(\xi-\hat{\omega})
        \prod_ke^{-r_k}\int_{[0,1]^N}e^{r_kf(t)}e^{-2\pi i\omega_k\cdot t}dt\\
    &= \sum_{\omega\in\mathbb{Z}^{N^2}}S_x(\xi-\hat{\omega})
        \prod_k\int_{[0,1]^N}e^{-r_k(1-f(t))}e^{-2\pi i\omega_k\cdot t}dt\\
    \end{split}
\end{equation}

Note that in the above equation, the first term is missing in the taylor series for the exponential.  We believe this is due to an error earlier in the derivation, since to reproduce known special cases such as the iid equations, this term is required.

Let $\mathcal{F}_k(\omega)=\int_{[0,1]^N}e^{-r_k(1-f(t))}e^{-2\pi i\omega\cdot t}dt$, and $\alpha_n=\sum_{k=1}^n\omega_{k}$.  This implies $\omega_1 = \alpha_1$, and $\omega_k = \alpha_k - \alpha_{k-1}$ for $k > 1$.  Also $\hat{\omega} = \alpha_N$.  Summing over all $\omega\in\mathbb{Z}^{N^2}$ can also be written as a sum over all $\alpha\in\mathbb{Z}^{N^2}$.

\begin{equation}
    \begin{split}
    S_y(\xi)
    &= \sum_{\omega\in\mathbb{Z}^{N^2}}S_x(\xi-\hat{\omega})
        \prod_k\mathcal{F}_k(\omega_k)\\
    &= \sum_{\alpha\in\mathbb{Z}^{N^2}}S_x(\xi-\alpha_N)
        \mathcal{F}_k(\alpha_1)
        \prod_{k=2}^N\mathcal{F}_k(\alpha_k-\alpha_{k-1})\\
    &= \sum_{\alpha_N\in\mathbb{Z}^N}S_x(\xi-\alpha_N)
        \sum_{\alpha_{N-1}\in\mathbb{Z}^N}\mathcal{F}_{N}(\alpha_N - \alpha_{N-1})
        \dots
        \sum_{\alpha_{1}\in\mathbb{Z}^N}\mathcal{F}_2(\alpha_2-\alpha_1)\mathcal{F}_1(\alpha_1)\\
    &= \sum_{\alpha_N\in\mathbb{Z}^N}S_x(\xi-\alpha_N)
        \left(\mathcal{F}_{N}*\dots*\mathcal{F}_1\right)(\alpha_N)\\
    \end{split}
\end{equation}

Using the convolution theorem for Fourier series coefficients:

\begin{equation}
    \begin{split}
    S_y(\xi)
    &= \sum_{\alpha_N\in\mathbb{Z}^N}S_x(\xi-\alpha_N)
        \int_{[0,1]^N}e^{-r(1-f(t))}e^{-2\pi i\alpha_N\cdot t}dt\\
    \end{split}
\end{equation}

Where $r=\sum_kr_k=4\pi^2\sigma_\varepsilon^2||\xi-\alpha_N||^2$.

To determine the structure of $f(t)$, we can analyze the autocorrelation of $\varepsilon(t)$:

\begin{equation}
    \begin{split}
    R_\varepsilon(t)
    &= \Ex[\varepsilon]{\varepsilon(\tau)\varepsilon(t+\tau)} = \Ex[u,v\sim p(u,v|t)]{uv} = \int_{\mathbb{R}^{2N}}uvp(u,v|t)dudv\\
    &= \int_{\mathbb{R}^{2N}}uv\int_{[0,1]^N}\prod_k\gamma_t(\rho_k)\mathcal{N}\left(u_k,v_k\left|0,
        \begin{bmatrix}\sigma_\varepsilon^2 & \rho_k\sigma_\varepsilon^2 \\ \rho_k\sigma_\varepsilon^2 & \sigma_\varepsilon^2\end{bmatrix}\right.\right)d\rho dudv\\
    &= \int_{[0,1]^N}\int_{\mathbb{R}^{2N}}\prod_k2u_kv_k\gamma_t(\rho_k)
        \mathcal{N}\left(u_k+v_k|0,2\sigma_\varepsilon^2\left(1+\rho_k\right)\right)
        \mathcal{N}\left(u_k-v_k|0,2\sigma_\varepsilon^2\left(1-\rho_k\right)\right)dudvd\rho\\
    &= \int_{[0,1]^N}\int_{\mathbb{R}^{2N}}\prod_k\frac{x_k+y_k}{2}\frac{x_k-y_k}{2}\gamma_t(\rho_k)
        \mathcal{N}\left(x_k|0,2\sigma_\varepsilon^2\left(1+\rho_k\right)\right)
        \mathcal{N}\left(y_k|0,2\sigma_\varepsilon^2\left(1-\rho_k\right)\right)dxdyd\rho\\
    &= \left(\int_0^1\frac{\gamma_t(\rho)}{4}\int_{\mathbb{R}^2}(x^2-y^2)
        \mathcal{N}\left(x|0,2\sigma_\varepsilon^2\left(1+\rho\right)\right)
        \mathcal{N}\left(y|0,2\sigma_\varepsilon^2\left(1-\rho\right)\right)dxdyd\rho\right)^N\\
    &= \left(\int_0^1\frac{\gamma_t(\rho)}{4}\left(
        \int_{-\infty}^\infty x^2\mathcal{N}\left(x|0,2\sigma_\varepsilon^2\left(1+\rho\right)\right)dx-
        \int_{-\infty}^\infty y^2\mathcal{N}\left(y|0,2\sigma_\varepsilon^2\left(1-\rho\right)\right)dy
        \right)d\rho\right)^N\\
    &= \left(\int_0^1\frac{\gamma_t(\rho)}{4}\left(
        2\sigma_\varepsilon^2\left(1+\rho\right)-2\sigma_\varepsilon^2\left(1-\rho\right)
        \right)d\rho\right)^N\\
    &= \left(\sigma_\varepsilon^2\int_0^1\rho\gamma_t(\rho)d\rho\right)^N = \sigma_\varepsilon^{2N}\mu_\gamma^N(t)\\
    \end{split}
\end{equation}

This implies that the mean of $\gamma_t(\rho)$ over $\rho$ as a function of $t$ is the normalized autocorrelation of the random field $\varepsilon(t)$, ie. $\mu_\gamma^N(t)=\rho_\varepsilon^N(t)$.  Using this result with the definition of $\gamma_t(\rho) = \delta(\rho-f(t))$ for the joint gaussian constraint on $\varepsilon(t)$, it follows that $f(t)=\rho_\varepsilon(t)$.

\section{Generalization of IID Analysis}
\label{app:iid-from-cont}

If discretization effects are ignored,

\begin{equation}
\begin{split}
    \sigma_y^2
    &=\int_{\mathbb{R}^N}S_y(\xi)d\xi\\
    &=\int_{\mathbb{R}^N}\sum_{\alpha\in\mathbb{Z}^N}S_x(\xi-\alpha)\int_{[0,1]^N}e^{-4\pi^2\sigma_\varepsilon^2||\xi-\alpha||^2\left(1-\rho_\varepsilon(t)\right)}e^{-2\pi i\alpha\cdot t}dtd\xi\\
    &=\sum_{\alpha\in\mathbb{Z}^N}\int_{\mathbb{R}^N}S_x(\xi-\alpha)\int_{[0,1]^N}e^{-4\pi^2\sigma_\varepsilon^2||\xi-\alpha||^2\left(1-\rho_\varepsilon(t)\right)}e^{-2\pi i\alpha\cdot t}dtd\xi\\
    &=\sum_{\alpha\in\mathbb{Z}^N}\int_{\mathbb{R}^N}S_x(\xi)\int_{[0,1]^N}e^{-4\pi^2\sigma_\varepsilon^2||\xi||^2\left(1-\rho_\varepsilon(t)\right)}e^{-2\pi i\alpha\cdot t}dtd\xi\\
    &=\int_{\mathbb{R}^N}S_x(\xi)\int_{[0,1]^N}e^{-4\pi^2\sigma_\varepsilon^2||\xi||^2\left(1-\rho_\varepsilon(t)\right)}\sum_{\alpha\in\mathbb{Z}^N}e^{-2\pi i\alpha\cdot t}dtd\xi\\
    &=\int_{\mathbb{R}^N}S_x(\xi)\int_{[0,1]^N}e^{-4\pi^2\sigma_\varepsilon^2||\xi||^2\left(1-\rho_\varepsilon(t)\right)}\sum_{\alpha\in\mathbb{Z}^N}\delta(t-\alpha)dtd\xi\\
    &=\int_{\mathbb{R}^N}S_x(\xi)\sum_{\alpha\in\mathbb{Z}^N}\int_{[0,1]^N}e^{-4\pi^2\sigma_\varepsilon^2||\xi||^2\left(1-\rho_\varepsilon(t)\right)}\delta(t-\alpha)dtd\xi\\
    &=\int_{\mathbb{R}^N}S_x(\xi)e^{-4\pi^2\sigma_\varepsilon^2||\xi||^2\left(1-\rho_\varepsilon(0)\right)}d\xi\\
    &=\int_{\mathbb{R}^2}S_x(\xi)d\xi\\
    &=\sigma_x^2
\end{split}
\end{equation}

Therefore, any deviation from this equivalence can be attributed to discretization effects.  To show this explicitly, we consider the input signal $x(t)\to x(t)*\varphi(t)$.  The effect on the power spectral density from this change would be $S_x(\xi)\to S_x(\xi)\phi(\xi)^2$, where $\phi(\xi)=\mathcal{F}\{\varphi(t)\}(\xi)$.

\begin{equation}
\begin{split}
    R_y(t)
    &=\int_{\mathbb{R}^N}S_y(\xi)e^{2\pi i\xi\cdot t}d\xi\\
    &=\int_{\mathbb{R}^N}\sum_{\alpha\in\mathbb{Z}^N}S_x(\xi-\alpha)\phi(\xi-\alpha)^2\int_{[0,1]^N}e^{-4\pi^2\sigma_\varepsilon^2||\xi-\alpha||^2\left(1-\rho_\varepsilon(\tau)\right)}e^{-2\pi i\alpha\cdot\tau}d\tau e^{2\pi i\xi\cdot t}d\xi\\
    &=\sum_{\alpha\in\mathbb{Z}^N}\int_{\mathbb{R}^N}S_x(\xi-\alpha)\phi(\xi-\alpha)^2\int_{[0,1]^N}e^{-4\pi^2\sigma_\varepsilon^2||\xi-\alpha||^2\left(1-\rho_\varepsilon(\tau)\right)}e^{-2\pi i\alpha\cdot\tau}d\tau e^{2\pi i\xi\cdot t}d\xi\\
    &=\sum_{\alpha\in\mathbb{Z}^N}\int_{\mathbb{R}^N}S_x(\xi)\phi(\xi)^2\int_{[0,1]^N}e^{-4\pi^2\sigma_\varepsilon^2||\xi||^2\left(1-\rho_\varepsilon(\tau)\right)}e^{-2\pi i\alpha\cdot\tau}d\tau e^{2\pi i\xi\cdot t}e^{2\pi i\alpha\cdot t}d\xi\\
    &=\int_{\mathbb{R}^N}S_x(\xi)\phi(\xi)^2\int_{[0,1]^N}e^{-4\pi^2\sigma_\varepsilon^2||\xi||^2\left(1-\rho_\varepsilon(\tau)\right)}\sum_{\alpha\in\mathbb{Z}^N}e^{-2\pi i\alpha\cdot(\tau-t)}d\tau e^{2\pi i\xi\cdot t}d\xi\\
    &=\int_{\mathbb{R}^N}S_x(\xi)\phi(\xi)^2\int_{[0,1]^N}e^{-4\pi^2\sigma_\varepsilon^2||\xi||^2\left(1-\rho_\varepsilon(\tau)\right)}\sum_{\alpha\in\mathbb{Z}^N}\delta(\tau-t-\alpha)d\tau e^{2\pi i\xi\cdot t}d\xi\\
    &=\int_{\mathbb{R}^N}S_x(\xi)\phi(\xi)^2\sum_{\alpha\in\mathbb{Z}^N}\int_{[0,1]^N}e^{-4\pi^2\sigma_\varepsilon^2||\xi||^2\left(1-\rho_\varepsilon(\tau)\right)}\delta(\tau-t-\alpha)d\tau e^{2\pi i\xi\cdot t}d\xi\\
\end{split}
\end{equation}

Assuming $\varepsilon$ and $x$ are white noise processes,

\begin{equation}
\begin{split}
    R_y(t)
    &=\int_{\mathbb{R}^N}\sigma_x^2\phi(\xi)^2e^{-4\pi^2\sigma_\varepsilon^2||\xi||^2\left(1-\rho_\varepsilon(\tau-\lfloor\tau\rfloor)\right)}e^{2\pi i\xi\cdot t}d\xi\\
    &=\int_{\mathbb{R}^N}\sigma_x^2\phi(\xi)^2e^{-4\pi^2\sigma_\varepsilon^2||\xi||^2}e^{2\pi i\xi\cdot t}d\xi\\
    &=\sigma_x^2\varphi(t)*\varphi(t)*\mathcal{N}(t;0,\sigma_\varepsilon^2\mathbf{I})
\end{split}
\end{equation}

Taking $\sigma_y^2=R_y(0)$,

\begin{equation}
    \sigma_y^2=\sigma_x^2\Ex{\varphi(t)*\varphi(t)}^d=\sigma_x^2\Phi_{2,1}^d
\end{equation}

Note that this analysis needed to assume that both $\varepsilon$ and $x$ are white noise processes to recover the structure of $\Phi_{2,1}$ present in the IID analysis.

\section{Unwarping}

Another way to ensure spatial consistency while using warping is to use the backwards pass of a warping block as it's own forward block.  It is clear from orthogonality analysis presented in the paper that this will result in an operator which given the same $\varepsilon$ as a warping block, will undo any spatial transformations.  As such, we call these unwarping blocks. We did not observe major empirical benefits from this structure, so the below analysis is merely for completeness.

Using the unwarping definition of the forward pass,

\begin{equation}
    y_{n\mathbf{v}} = \sum_\mathbf{u}x_{n\mathbf{u}}\prod_{k}\varphi\left(\mathbf{u}_k-\mathbf{v}_k+\varepsilon_{k\mathbf{u}}\right)
\end{equation}

The gradient with respect to $x$ can be determined as follows,

\begin{equation}
\begin{split}
    \nabla_{x} \mathcal{L}_{i\mathbf{w}}
    &= \partial_{x_{i\mathbf{w}}} \mathcal{L}
     = \sum_{n,\mathbf{v}} \nabla_y\mathcal{L}_{n\mathbf{v}}\partial_{x_{i\mathbf{w}}} y_{n\mathbf{v}}\\
    &= \sum_{n,\mathbf{v}} \nabla_y\mathcal{L}_{n\mathbf{v}}\sum_\mathbf{u}x_{n\mathbf{u}}\prod_{k}\varphi\left(\mathbf{u}_k-\mathbf{v}_k+\varepsilon_{k\mathbf{u}}\right)\\
    &= \sum_{n,\mathbf{v}} \nabla_y\mathcal{L}_{nv}\sum_\mathbf{u}\delta_{in}\delta_{\mathbf{wu}}\prod_{k}\varphi\left(\mathbf{u}_k-\mathbf{v}_k+\varepsilon_{k\mathbf{u}}\right)\\
    &= \sum_{\mathbf{v}} \nabla_y\mathcal{L}_{i\mathbf{v}}\prod_{k}\varphi\left(\mathbf{u}_k-\mathbf{v}_k+\varepsilon_{k\mathbf{u}}\right)\\
\end{split}
\end{equation}

And the gradient with respect to $\varepsilon$ can be determined as follows,

\begin{equation}
\begin{split}
    \nabla_\varepsilon \mathcal{L}_{i\mathbf{w}}
    &= \partial_{\varepsilon_{i\mathbf{w}}} \mathcal{L}
     = \sum_{n,\mathbf{v}} \nabla_y\mathcal{L}_{n\mathbf{v}}\partial_{\varepsilon_{iw}} y_{n\mathbf{v}}\\
    &= \sum_{n,\mathbf{v}} \nabla_y\mathcal{L}_{n\mathbf{v}}\sum_\mathbf{u}x_{n\mathbf{u}}\prod_{k}\varphi\left(\mathbf{u}_k-\mathbf{v}_k+\varepsilon_{k\mathbf{u}}\right)\\
    &= \sum_{n,\mathbf{v}} \nabla_y\mathcal{L}_{n\mathbf{v}}\sum_\mathbf{u}x_{n\mathbf{u}} \partial_{\varepsilon_{i\mathbf{w}}}\varphi\left(\mathbf{u}_i-\mathbf{v}_i+\varepsilon_{i\mathbf{u}}\right) \prod_{k\neq i}\varphi\left(\mathbf{u}_k-\mathbf{v}_k+\varepsilon_{k\mathbf{u}}\right)\\
    &= \sum_{n,\mathbf{v}} \nabla_y\mathcal{L}_{n\mathbf{v}}\sum_\mathbf{u}x_{n\mathbf{u}} \varphi^\prime\left(\mathbf{u}_i-\mathbf{v}_i+\varepsilon_{i\mathbf{u}}\right) \delta_{uw} \prod_{k\neq i}\varphi\left(\mathbf{u}_k-\mathbf{v}_k+\varepsilon_{k\mathbf{u}}\right)\\
    &= \sum_{n,\mathbf{v}} \nabla_y\mathcal{L}_{n\mathbf{v}}x_{n\mathbf{w}} \varphi^\prime\left(\mathbf{w}_i-\mathbf{v}_i+\varepsilon_{i\mathbf{w}}\right) \prod_{k\neq i}\varphi\left(\mathbf{w}_k-\mathbf{v}_k+\varepsilon_{k\mathbf{w}}\right)\\
    &= \sum_{n} x_{n\mathbf{w}} \sum_{\mathbf{v}} \nabla_y\mathcal{L}_{n\mathbf{v}}\varphi^\prime\left(\mathbf{w}_i-\mathbf{v}_i+\varepsilon_{i\mathbf{w}}\right) \prod_{k\neq i}\varphi\left(\mathbf{w}_k-\mathbf{v}_k+\varepsilon_{k\mathbf{w}}\right)\\
\end{split}
\end{equation}

\section{Warping Examples}

\begin{figure}
    \centering
    \includegraphics[width=0.85\textwidth]{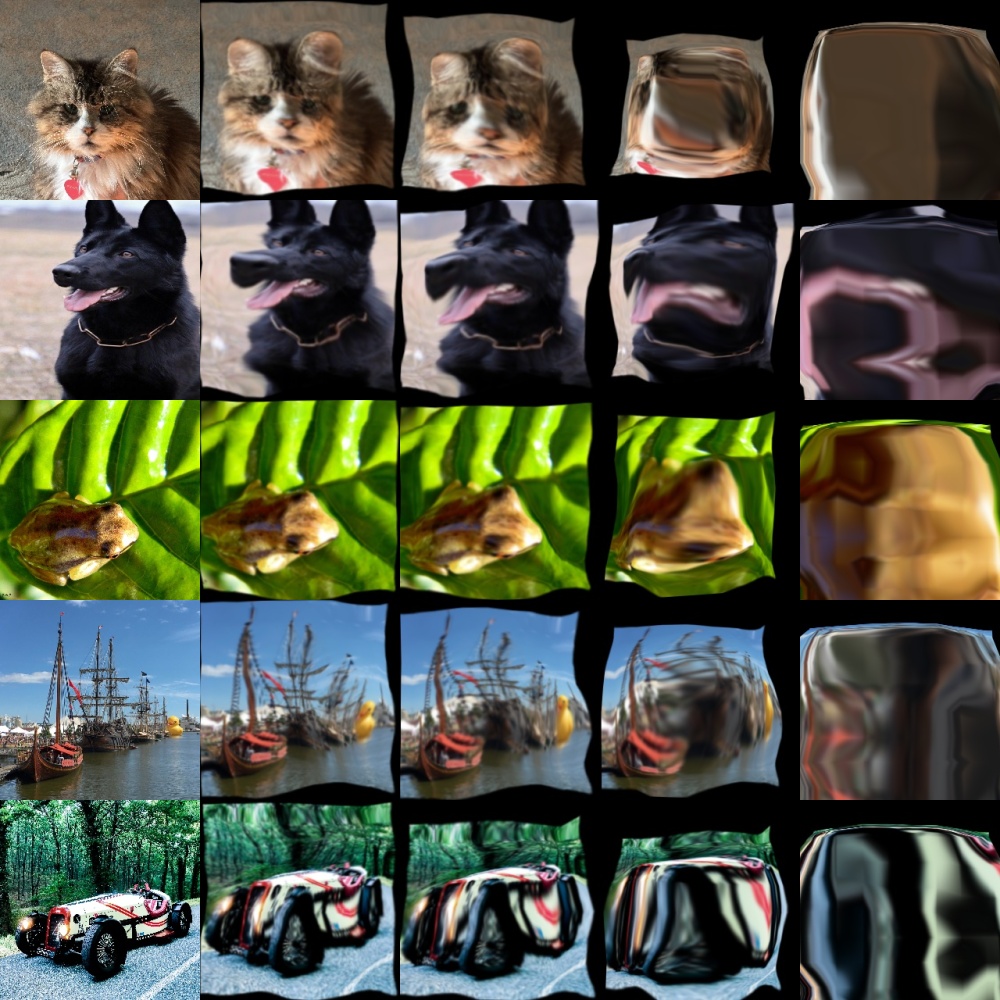}
    \caption{Sample images with warping throughout a Resnet-56 warping network.}
    \label{fig:warping-grid}
\end{figure}

To show the qualitative characteristics of the warping transformation from a fully learned model, we have generated \ref{fig:warping-grid}. The model used to determine the warpings is the same as the highest accuracy model from the warping ablation results (Resnet-56 based architecture), which was trained on Cifar-10. Rather than directly using images from the data set, we chose higher resolution images to better demonstrate warping, though we used a downscaled version of each of the images as input to the model. From these input images, we then extracted the warping offsets $\varepsilon$, upscaled them, and iteratively applied them to the original image. It should be noted that the model itself also utilized convolutions and other transformations, so this visualization is merely an approximation of only the warping components as applied to the input directly. Additionally, many instances of the warping operation are not included in the image, especially warps from early in the network, as many of these layers collapse during training and do not transform the data in a meaningful way (as mentioned in the bifurcation analysis). The chosen images were gathered after the transformations from cumulatively 48, 50, 52, and 54 (last layer) warping layers, with the first image being the input. From viewing the results, we can see where the model focuses, and the ability of warping to dynamically transform the data to localize to areas of highest information.

\end{document}